\documentclass[compsoc,  onecolumn]{IEEEtran}

\usepackage[center]{caption}
\usepackage[dvipsnames]{xcolor}

\usepackage{cite,url}
\usepackage{amsmath}
\usepackage{amssymb}
\usepackage{mathrsfs}
\usepackage{nameref}
\usepackage[inline]{enumitem}
\usepackage{soul}
\usepackage[colorlinks=true,linkcolor=black,anchorcolor=black,citecolor=black,filecolor=black,menucolor=black,runcolor=black,urlcolor=black]{hyperref}

\usepackage{verbatim} 
\usepackage{makecell} 

\usepackage{pifont}
\newcommand{\xmark}{\ding{55}}
\newcommand{\cmark}{\ding{51}}
\definecolor{cadmiumgreen}{rgb}{0.0, 0.42, 0.24} 
\usepackage{bm} 

\usepackage{xr}
\usepackage{todonotes}
\AtBeginDocument{%
  \providecommand\BibTeX{{%
    \normalfont B\kern-0.5em{\scshape i\kern-0.25em b}\kern-0.8em\TeX}}}

\author{Denis Kleyko, Dmitri A. Rachkovskij, Evgeny Osipov, and Abbas Rahimi\\ 
\thanks{We would like to thank three reviewers, the editors, and Pentti Kanerva for their insightful feedback as well as Linda Rudin for the careful proofreading that contributed to the final shape of the survey. The work of DK was supported by the European Union's Horizon 2020 Programme under the Marie Skłodowska-Curie Individual Fellowship Grant (839179). The work of DK  was also supported in part by AFOSR FA9550-19-1-0241 and Intel's THWAI program.
The work of DAR was supported in part by the National Academy of Sciences of Ukraine (grant no. 0120U000122, 0121U000016, and 0117U002286), the Ministry of Education and Science of Ukraine (grant no. 0121U000228 and 0122U000818), and the Swedish Foundation for Strategic Research (SSF, grant no. UKR22-0024).
\textit{Denis Kleyko and Dmitri A. Rachkovskij contributed equally to this work.}
}
\thanks{D. Kleyko is with the Redwood Center for Theoretical Neuroscience at the University of California, Berkeley, CA 94720, USA and also with the Intelligent Systems Lab at Research Institutes of Sweden, 16440 Kista, Sweden. \mbox{E-mail}: \mbox{denkle@berkeley.edu}
}
\thanks{D. A. Rachkovskij is with International Research and Training Center for Information Technologies and Systems, 03680 Kiev, Ukraine. 
\mbox{E-mail}: \mbox{dar@infrm.kiev.ua}
 }

\thanks{E. Osipov is with the Department of Computer  Science Electrical and Space Engineering, Lule\aa{} University of Technology, 97187 Lule\aa{}, Sweden. \mbox{E-mail}: \mbox{evgeny.osipov@ltu.se}
}
\thanks{A. Rahimi is with IBM Research, 8803 Zurich,  Switzerland. \mbox{E-mail}: \mbox{abr@zurich.ibm.com}
}
}

\begin{document}

\title{
A Survey on Hyperdimensional Computing \\
aka Vector Symbolic Architectures, Part I: \\
Models and Data Transformations
}

\maketitle

\begin{abstract}
This two-part comprehensive survey is devoted to a computing framework most commonly known under the names Hyperdimensional Computing and Vector Symbolic Architectures (HDC/VSA).
Both names refer to a  family of computational models that use high-dimensional distributed representations and rely on the algebraic properties of their key operations to incorporate the advantages of structured symbolic representations  and vector distributed representations.
Notable models in the HDC/VSA family are Tensor Product Representations, Holographic Reduced Representations, Multiply-Add-Permute, Binary Spatter Codes, and Sparse Binary Distributed Representations but there are other models too. 
HDC/VSA is a highly interdisciplinary field with connections to computer science, electrical engineering, artificial intelligence, mathematics, and cognitive science. This fact makes it challenging to create a thorough overview of the field. However, due to a surge of new researchers joining the field in recent years, the necessity for a comprehensive survey of the field has become extremely important. Therefore, amongst other aspects of the field, this Part I surveys  important aspects such as: known computational models of HDC/VSA and transformations of various input data types to high-dimensional distributed representations. 
Part~II of this survey~\cite{KleykoSurveyVSA2021Part2} is devoted to
applications, cognitive computing and architectures, as well as directions for future work.
The survey is written to be useful for both newcomers and practitioners.
\end{abstract}

\begin{IEEEkeywords}
Artificial Intelligence,
Machine learning, 
Distributed representations,
Data structures,
Hyperdimensional Computing,
Vector Symbolic Architectures, 
Holographic Reduced Representations,
Tensor Product Representations,
Matrix Binding of Additive Terms,
Binary Spatter Codes,
Multiply-Add-Permute,
Sparse Binary Distributed Representations,
Sparse Block Codes,
Modular Composite Representations,
Geometric Analogue of Holographic Reduced Representations
\end{IEEEkeywords}

\tableofcontents
\newpage

\section{Introduction}

The two main approaches to Artificial Intelligence (AI) are symbolic and connectionist. The  symbolic approach represents information via symbols and their relations. Symbolic AI (the alternative term is Good Old-Fashioned AI, GOFAI) solves problems or infers new knowledge through the processing of these symbols. 
In the alternative connectionist approach, information is processed in a network of simple computational units often called neurons, so another name for the connectionist approach is artificial neural networks. This article presents a survey of a research field that originated at the intersection of GOFAI and connectionism, which is known under the names Hyperdimensional Computing, HDC (term introduced in~\cite{KanervaHyperdimensional2009}) and Vector Symbolic Architectures, VSA (term introduced in~\cite{GaylerJackendoff2003}).

In order to be consistent and to avoid possible confusion amongst the researchers outside the field, we will use the joint name HDC/VSA when referring to the field.
It is also worth pointing out that probably the most influential and well-known (at least in the machine learning domain) HDC/VSA model is Holographic Reduced Representations~\cite{PlateHolographic2003} and, therefore, this term is also used when referring to the field. 
For the reason of consistency, however, we use HDC/VSA as a general term while referring to Holographic Reduced Representations when discussing this particular model.  
HDC/VSA is the umbrella term for a  family of computational models that rely on mathematical properties of high-dimensional  vector spaces and use high-dimensional distributed representations called hypervectors (HVs\footnote{
Another term to refer to HVs, which is commonly used in the cognitive science literature, is ``semantic pointers''~\cite{BlouwConcepts2016}.}) for structured (``symbolic'') representation of data while maintaining the advantages of connectionist vector distributed representations. This opens  
a promising way to build AI systems~\cite{LevyNewBuilding2008}.

For a long time HDC/VSA did not gain much attention from the AI community.
Recently, the situation has, however, begun to change and right now HDC/VSA is picking up momentum. 
We attribute this to a combination of several factors such as dissemination efforts from the members of the research community\footnote{HDC/VSA Web portal. [Online.] Available: \url{https://www.hd-computing.com}}\footnote{VSAONLINE. Webinar series. [Online.] Available: \url{https://sites.google.com/ltu.se/vsaonline}}, several successful engineering applications in the machine learning domain (e.g.,~\cite{SahlgrenRIIntroduction2005, RahimiBiosignal2019}) and cognitive architectures~\cite{EliasmithSPAUN2012, RachkovskijWorldModel2013}.
The main driving force behind the current interest is the
global trend of searching for computing paradigms alternative to the conventional (von Neumann) one, such as neuromorphic and nanoscalable  computing, where HDC/VSA is  expected to play an important role (see~\cite{KleykoComputingParadigm2021} and references therein for perspective). 

The major problem for researchers new to the field  is that the previous work on HDC/VSA is spread across many venues and disciplines and cannot be tracked easily. Thus, understanding the state-of-the-art of the field is not trivial.
Therefore, in this article we survey HDC/VSA with the aim of providing the broad coverage of the field, which is currently missing.  
While the survey is written to be accessible to a wider audience, it should not be considered as ``the easiest entry point''. 
For anyone who has not yet been exposed to HDC/VSA, before reading this survey, we highly recommend starting with three tutorial-like introductory articles~\cite{KanervaHyperdimensional2009, KanervaComputing2019}, and~\cite{NeubertRobotics2019}. 
The former two  provide a solid introduction and motivation behind the field while the latter focuses on introducing HDC/VSA within the context of a particular application domain of robotics.
Someone who is looking for a very concise high-level introduction to the field without too many technical details might consult~\cite{KanervaWords2014}.
Finally, there is a book~\cite{PlateHolographic2003} that provides a comprehensive treatment of fundamentals of two particular HDC/VSA models (see Sections~\ref{sec:vsa:hrr} and~\ref{sec:vsa:fhrr}).
While~\cite{PlateHolographic2003} focused on the two specific models, many of the aspects presented there apply to HDC/VSA in general.

\begin{table}[tb]
\renewcommand{\arraystretch}{1.0}
\caption{A qualitative assessment of existing HDC/VSA literature with elements of survey.}
\label{table:position}
    \begin{center}
    \begin{tabular}{c c c c c c c c c c } 
     & \rotatebox[origin=c]{90}{\makecell{Representation \\  
types}} 
& \rotatebox[origin=c]{90}{\makecell{Connectionism \\ challenges}}   
& \rotatebox[origin=c]{90}{\makecell{The HDC/VSA \\ models }}   
& \rotatebox[origin=c]{90}{\makecell{ Capacity of \\ hypervectors}}   
& \rotatebox[origin=c]{90}{\makecell{Hypervectors for \\ symbols and sets}} 
& \rotatebox[origin=c]{90}{\makecell{Hypervectors for \\  numeric data }}   
& \rotatebox[origin=c]{90}{\makecell{Hypervectors for \\  sequences }} 
& \rotatebox[origin=c]{90}{\makecell{Hypervectors for \\  2D images }} 
& \rotatebox[origin=c]{90}{\makecell{Hypervectors for \\  graphs }} 

\\ \hline

     \cite{PlateCommon1997} & $\bm{\pm}$ & $\bm{\pm}$ & $\bm{\pm}$ & {\color{red} \xmark} & {\color{cadmiumgreen} \cmark} & {\color{red} \xmark}  & $\bm{\pm}$ & {\color{red} \xmark} & $\bm{\pm}$   \\  \hline 
     
     \cite{KanervaHyperdimensional2009} & $\bm{\pm}$ & $\bm{\pm}$  & $\bm{\pm}$ & {\color{red} \xmark} & {\color{cadmiumgreen} \cmark}  & {\color{red} \xmark} & $\bm{\pm}$ & {\color{red} \xmark} & {\color{red} \xmark}  \\  \hline 
     
     \cite{RahimiNanoscalable2017}& {\color{red} \xmark} & {\color{red} \xmark} & $\bm{\pm}$ & $\bm{\pm}$ & {\color{cadmiumgreen} \cmark} & {\color{red} \xmark} & $\bm{\pm}$ & {\color{red} \xmark} & {\color{red} \xmark}   \\  \hline 
     
     \cite{RahimiBiosignal2019}& {\color{red} \xmark} & {\color{red} \xmark} &  {\color{red} \xmark} & {\color{red} \xmark} & {\color{cadmiumgreen} \cmark} & $\bm{\pm}$ & $\bm{\pm}$ & {\color{red} \xmark} & {\color{red} \xmark} \\  \hline      
     
     \cite{NeubertRobotics2019}& {\color{red} \xmark} & {\color{red} \xmark} &  $\bm{\pm}$ & {\color{red} \xmark} & {\color{cadmiumgreen} \cmark} & $\bm{\pm}$  & $\bm{\pm}$ & {\color{red} \xmark} & {\color{red} \xmark}  \\  \hline     
     
     \cite{GeClassificationReview2020} & {\color{red} \xmark} & {\color{red} \xmark} &  {\color{red} \xmark} & {\color{red} \xmark} & {\color{cadmiumgreen} \cmark} & {\color{red} \xmark} & $\bm{\pm}$ & {\color{red} \xmark} & {\color{red} \xmark}  \\  \hline 
     
     \cite{SchlegelVSAComparison2020} & {\color{red} \xmark} & {\color{red} \xmark}  & $\bm{\pm}$ & $\bm{\pm}$ & {\color{cadmiumgreen} \cmark}  & {\color{red} \xmark} & $\bm{\pm}$ & {\color{red} \xmark} & {\color{red} \xmark}  \\  \hline    
     
     \cite{KleykoComputingParadigm2021} & {\color{red} \xmark} & {\color{red} \xmark} &  {\color{red} \xmark}  & {\color{red} \xmark} & {\color{cadmiumgreen} \cmark} & {\color{red} \xmark} & {\color{cadmiumgreen} \cmark} & {\color{red} \xmark} & $\bm{\pm}$    \\  \hline 
     
     \cite{hassan2021hyper} & {\color{red} \xmark} & {\color{red} \xmark} &  {\color{red} \xmark} & {\color{red} \xmark} & {\color{cadmiumgreen} \cmark}  & {\color{red} \xmark}  & $\bm{\pm}$ & {\color{red} \xmark} & {\color{red} \xmark}  \\ \hline    
     \hline  
     
    
    \makecell{This survey, Part I \\ Section \#} & \ref{sec:motivation:types} & \ref{sec:challenges} &  \ref{sec:vsa:frameworks} & \ref{sec:capacity} & \ref{sec:symb:sets} & \ref{sec:scalars:vectors} & \ref{sec:sequences} & \ref{sec:2Dimages} & \ref{sect:graphs}
    \\ \hline        
     
     \end{tabular}
    \end{center}
\end{table}

To our knowledge, there have been no previous attempts to make a comprehensive survey of HDC/VSA but there are articles that overview particular topics of HDC/VSA. 
Table~\ref{table:position} contrasts the coverage of this survey with those previous articles (listed chronologically). We use $\bm{\pm}$ to indicate that an article partially addressed a particular topic, but either new results have been reported since then or not all related work was covered.

Part~I of this survey has the following structure. 
In Section~\ref{sec:hdcvsa}, we introduce the motivation behind HDC/VSA, their basic notions, and summarize currently known HDC/VSA models.
Section~\ref{sec:data:to:HV} presents transformation of various data types to HVs.
Discussion and conclusions follow in Sections~\ref{sec:disc} and~\ref{sec:conc}, respectively.

Part~II of this survey~\cite{KleykoSurveyVSA2021Part2}) will cover existing applications and the use of HDC/VSA in cognitive architectures. 

Finally, due to the space limitations, there are topics that remain outside the scope of this survey. 
These topics include connections between HDC/VSA and other research fields as well as hardware implementations of HDC/VSA.
We plan to cover these issues in a separate work.

\section{Hyperdimensional Computing aka Vector Symbolic Architectures}
\label{sec:hdcvsa}

In this section, we describe the motivation that led to the development of early HDC/VSA models (Section~\ref{sec:motivation}), list 
 their components (Section~\ref{sec:VSA:structure-sensitive}), 
overview existing HDC/VSA models (Section~\ref{sec:vsa:frameworks}) and discuss the information capacity of HVs (Section~\ref{sec:capacity}).

\subsection{Motivation and basic notions}
\label{sec:motivation}

The ideas relevant for HDC/VSA already appeared in the late 1980s and early 1990s~\cite{KanervaSDM1988, MizrajiContext1989, SmolenskyTensor1990, RachkovskijAudio1990, PlateAlgebra1991, KussulAPNN1991}.
In this section, we first review the types of representations together with their advantages and disadvantages. Then we introduce some of the challenges posed to distributed representations, which turned out to be motivating factors for inspiring the development of HDC/VSA.

\subsubsection{Types of representation}
\label{sec:motivation:types}

\paragraph{Symbolic representations}
Symbolic representations~\cite{NewellSymbols1976,HarnadSymbol1990} are natural for humans and widely used in computers. 
In symbolic representations, each item or object is represented by a symbol. 

Here, by objects we refer to items of various nature and complexity,
such as features, relations, physical objects, scenes, their classes, etc. More complex symbolic representations can be composed from the simpler ones. 

Symbolic representations naturally possess a combinatorial structure that allows producing indefinitely many propositions/symbolic expressions (see Section~\ref{sec:challenges:Fodor} below). This is achieved by composition using rules or programs as demonstrated by, e.g., Turing Machines. This process is, of course, limited by memory size that can, however, be expanded without changing the computational structure of a system. 
A vivid example of a symbolic representation/system is a natural language, where a plethora of words is composed from a small alphabet of letters. In turn, a finite number of words is used to compose an infinite number of sentences and so on.

Symbolic representations have all-or-none explicit similarity: the same symbols have maximal similarity, whereas different symbols have zero similarity and are, therefore, called dissimilar. To process symbolic structures, one needs to follow edges and/or match vertices of underlying graphs to, e.g., reveal the whole structure or calculate the similarity between composite symbolic structures. Therefore, symbolic models usually have problems with scaling because similarity search and reasoning in such models require complex and sequential operations, which quickly become intractable.

In the context of brain-like computations, the downside of the conventional implementation of symbolic computations is that they require reliable hardware~\cite{WangCharacterizing2004}, since any error in computation might result in a fatal fault. In general, it is also unclear how symbolic representations and computations with them could be implemented in a biological tissue, especially when taking into account its unreliable nature.

\paragraph{Connectionist representations: localist and distributed}
\label{sec:connectionist:representations}
In this article, connectionism is used as an umbrella term for approaches related to neural networks and brain-like computations. 
Two main types of connectionist representations are distinguished: localist and distributed~\cite{GelderDistributed1999, ThorpeLocalized1998}.

Localist representations are akin to symbolic representations in that for each object there exists a single corresponding element in the implementation of the representation. 
Examples of localist representations are a single neuron (node) or a single vector component.

There is some evidence that localist representations might be used in the brain (so-called ``grandmother cells'')~\cite{QuirogaInvariantBrain2005}. In order to link localist representations, connections between components can be created, corresponding to pointers in symbolic representations.
However, constructing compositional structures that include combinations of already represented objects requires the allocation of (a potentially infinite number of) new additional elements and connections, which is neurobiologically questionable. For example, representing ``abc'', ``abd'', ``acd'', etc. requires introducing new elements for them, as well as connections between ``a'', ``b'', ``c'', ``d'' and so on. Also, localist representations share symbolic representations' drawbacks of lacking enough semantic basis, that may be considered as a lack of immediate explicit similarity between the representations. 
In other words, different neurons representing different objects are dissimilar and estimating object similarity requires additional processes.

Distributed representations\footnote{
Note that here we discuss not only distributed representations in the form of HVs formed with HDC/VSA but also distributed representations in general, including the ones used in early connectionist approaches. 
We refer to the latter as conventional connectionist representations.  
} were inspired by the idea of a  ``holographic'' representation as an alternative to the localist representation \cite{HintonDistributed1986, ThorpeLocalized1998, GelderDistributed1999, PlateDistributed2006}. They are attributed to a connectionist approach based on modeling the representation of information in the brain as ``distributed'' over many neurons. In distributed representations, the state of a set of neurons (of finite size) is modeled as a vector where each vector component represents a state of the particular neuron. %

Distributed representations are defined as a form of vector representations, where each object is represented by a subset of vector components, and each vector component can belong to representations of many objects. This concerns (fully distributed) representations of objects of various complexity, from elementary features or atomic objects to complex scenes/objects that are represented by (hierarchical) compositional structures. 

In distributed representations, the state of individual components of the representation cannot be interpreted without knowing the states of other components. In other words, in distributed representations the semantics of individual components of the representation are usually undefined, in distinction to localist representations.

In order to be useful in practice for engineering applications and cognitive modeling, distributed representations of similar objects should be similar (according to some similarity measure of the corresponding vector representations; see Section~\ref{sec:vsa:similarity}), thus addressing the semantic basis issue of symbolic and localist representations. 

Distributed representations should have the following attractive properties:
\begin{itemize}
\item High representational capacity. For example, if one object is represented by $M$ binary components of a $D$-dimensional vector, then the number of representable objects equals the number of combinations $\binom{D}{M}$, 
in contrast to $D / M$ for localist representations;

\item Explicit representation of similarity. Similar objects have similar representations that can be immediately compared by efficiently computable vector similarity measures (e.g., dot product, Minkowski distance, etc.); 
\item A rich semantic basis due to the immediate use of representations based on features and the possibility of representing the similarity of the features themselves in their vector representations; 
\item The possibility of using well-developed methods for processing vectors; 
\item For many types of distributed representations -- the ability to recover the original representations of objects; 
\item Ability to work in the presence of noise, malfunction, and uncertainty, in addition to neurobiological plausibility. 

\item Direct access to the representation of an object. Being a vector, a distributed representation of a compositional structure can be processed directly. This does not require tracing pointers as in symbolic representations or following connections between components as in localist representations; 

\item Unified format. Every object, whether atomic or composite, is represented by a vector and, hence, the implementation operates at the level of vectors without being explicitly aware of the complexity of what is represented.

\end{itemize}

In our opinion, the last two properties listed above are closely connected to the challenges posed to distributed representations in the late 1980s and early 1990s that we briefly discuss in the next section.

\subsubsection {Challenges for conventional connectionist representations}
\label{sec:challenges}

Let us consider several challenges faced by early distributed representations known as ``superposition catastrophe'', e.g.,~\cite{MalsburgAssemblies1986, RachkovskijBinding2001}. And, at a higher level, by demand for ``systematicity''~\cite{FodorCritical1988}, and much later, for fast compositionality~\cite{Jackendoff2002}. 
These challenges led to the necessity to make distributed representations ``structure-sensitive'' by introducing the ``binding'' operation (Section~\ref{sec:binding}).

\paragraph{ ``Superposition catastrophe''}
\label{sec:sup:catastrophe}
It was believed that connectionist representations cannot represent hierarchical compositional structures because of the superposition catastrophe, which manifestates itself in losing the information concerning object arrangements in structures, see~\cite{MalsburgAssemblies1986, RachkovskijBinding2001, RachkovskijStructures2001} for discussion and references. 

In the simplest case, let us activate binary localist representation elements corresponding to ``a'' \& ``b'', then to ``b'' \& ``c''. If we want to represent both ``a'' \& ``b'', and ``b'' \& ``c'' simultaneously, we activate all three representations: ``a'', ``b'', ``c''. Now, however, the information that ``a'' was with ``b'', and ``b'' was with ``c'' is lost. For example, the same ``a'', ``b'', ``c'' activation could be obtained by ``a'' \& ``c'', and single ``b''. The same situation occurs if ``a'', ``b'', ``c'' are represented by distributed patterns.

\paragraph{Fodor \& Pylyshyn criticisms of connectionism}
\label{sec:challenges:Fodor}
In~\cite{FodorCritical1988}, criticism of connectionism was concerned with the parallel distributed processing approach covered in~\cite{HintonDistributed1986}. Fodor and Pylyshyn claimed that connectionism lacks Productivity, Systematicity, Compositionality, and Inferential Coherence that are inherent to systems operating with symbolic representations. Their definitions of these intuitively appealing issues are rather vague and interrelated, and their criticism is constrained to the early particular connectionist model that the authors chose for their critique. 
Therefore, we restate these challenges as formulated in~\cite{PlateHolographic2003}: 
\begin{itemize}
\item Composition, decomposition, and manipulation: How are elements composed to form a structure, and how are elements extracted from a structure? Can the structures be manipulated using distributed representations? 
\item  Productivity: A few simple rules for composing elements can give rise to a huge variety of possible structures. Should a system be able to represent structures unlike any it has previously encountered, if they are composed of the same elements and relations? 
 \item Systematicity: Does the distributed representation allow processes to be sensitive to the structure of the objects? To what degree are the processes independent of the identity of elements in compositional structures? 
\end{itemize}

\paragraph{Challenges to connectionism posed by Jackendoff}
Four challenges to connectionism have been posed by Jackendoff~\cite{Jackendoff2002}, see also~\cite{GaylerJackendoff2003} for their HDC/VSA treatment. In principle, they are relevant to cognition but in particular they are related to language. The problem, in general, is how to neurally instantiate the rapid construction and transformation of the compositional structures. 

\begin{itemize}
    \item \textit{Challenge 1. The binding problem}: the observation that linguistic representations must use compositional representations, taking into account order and occurring combinations. For example, the same words in different order and combination are going to produce different sentences.
    \item \textit{Challenge 2. The problem of two}:  how are multiple instances of the same object instantiated? For example, how are the ``little star'' and the ``big star'' instantiated so that they are both stars, yet distinguishable? 
    
    \item \textit{Challenge 3. The problem of variables}: concerns typed variables. One should be able to represent templates or relations with variables (e.g., names of relations) and values (e.g., arguments of relations).
    
    \item \textit{Challenge 4. Binding in working and in long-term memories}: representations of the same binding should be identical in various types of memory. In other words, the challenge concerns the transparency of the boundary between a working memory and a long-term memory. It has been argued that linguistic tasks require the same structures to be instantiated in the working memory and the long-term memory and that the two instantiations should be functionally equivalent.

\end{itemize}

\subsubsection {Binding to address challenges of conventional connectionist representations}
\label{sec:motivation:binding}

The challenges presented above made it clear that an adequate representation of compositional structures requires preserving information about their grouping and order. 
For example, in symbolic representations, brackets and symbolic order can be used to achieve this. 
For the same purpose, some mechanism of binding (\`{a} la ``grouping brackets'') was needed in distributed representations.

We view the binding operation as a means to form  such a representation of an object that contains information about the context in which it was encountered.
So, any implementation of the binding operation is expected to involve some modification of the representation of the original object.
The context can essentially be anything, such as other homogeneous objects (e.g., data objects of the same type) or heterogeneous objects (e.g., data objects' positions, roles, etc.).
The binding operation, together with other operations, should provide a mechanism for constructing representations of compositional objects that reflect their similarity in a way that is useful for the problem being solved. For example, different combinations of the same objects should be represented differently.
Also, it is often required to support a recovery (of the representation) of the original composite object.

One of the approaches to binding in distributed representations is based on the temporal synchronization of constituent activations~\cite{MilnerShape1974,  MalsburgAssemblies1986, ShastriSynchrony1993, HummelDistributed1997}.
Although this mechanism may be useful on a single level of composition, its capabilities to represent compositional structures with multiple levels of hierarchy are questionable as it requires many time steps and complex ``orchestration'' to represent compositional structures. 
Another major problem is that such a temporal representation cannot be immediately stored in a long-term memory.

An alternative approach to binding, which eliminates these issues, is the so-called conjunctive coding approach used in HDC/VSA. Its predecessors were ``extra units'' considered by~\cite{HintonImplementing1981} to represent various combinations of active units of distributed patterns as well as outer products~\cite{MizrajiContext1989,SmolenskyTensor1990},  which, however, increased the dimensionality of representations (see details in Section~\ref{sec:framework:TPR}). 
In HDC/VSA, the binding operation does not change the dimensionality of distributed representations. Moreover, it does not require any training. 

To form ``rich'' compositional representations in HDC/VSA, both binding and superposition operations are used. Therefore, implementations of these operations need to maintain their properties when used together.
In particular, the binding operation should not be associative with respect to the superposition operation in order to overcome the superposition catastrophe (Section~\ref{sec:sup:catastrophe}).

Importantly, the schemes for forming distributed representations of compositional structures that exploit the binding and superposition operations produce distributed representations that are similar for similar objects (i.e., they take into account the similarity of object elements, their grouping, and order at various levels of hierarchy).

In summary, the main motivation for developing HDC/VSA was to combine the advantages of early distributed representations and those of symbolic representations, while avoiding their drawbacks, in pursuit for more efficient information processing and, ultimately, for better AI systems. One of the goals was to address the above challenges faced by conventional connectionist representations. The properties of HDC/VSA models introduced below allow addressing these challenges to a varying degree.

\subsection{Structure-sensitive distributed representations}
\label{sec:VSA:structure-sensitive}

\subsubsection{Atomic representations}
\label{sect:atom:repr}

When designing  an HDC/VSA-based system it is common to define a set of the most basic objects/items/entities/concepts/ symbols/scalars for the given problem and assign them HVs, which are referred to as atomic HVs.
The process of assigning atomic HVs is often referred to as mapping, projection, embedding, formation or transformation. 
To be consistent, we will use the term transformation.

For a given problem, we need to choose atomic representations of objects such that the similarity between the representations of objects corresponds to the properties we care about.\footnote{
It is assumed that the problem to be solved has a static similarity structure. There may be problems that require a dynamic similarity structure, but these have not, so far, been the subjects of extensive study.
}
HVs of other (compositional) objects are formed by the atomic HVs (see Section~\ref{sec:compositional}).
As follows from their name, atomic HVs are high-dimensional vectors. Values of HV components could be binary, real, or complex numbers and this list is not exhaustive, as we will see in Section~\ref{sec:vsa:frameworks}.

In the early days of HDC/VSA, most of the works were focused on symbolic problems. 
In the case of working with symbols one could easily imagine many problems where a reasonable assumption would be that symbols are not related at all. 
So their atomic HVs are generated at random and are considered dissimilar, i.e., their expected similarity value is considered to be ``zero''.
On the other hand,  there are many problems where assigning atomic HVs fully randomly does not lead to any useful behavior of the designed system, see Section~\ref{sec:scalars:vectors}.

\begin{figure}[t]
\centering
\includegraphics[width=1.0\columnwidth]{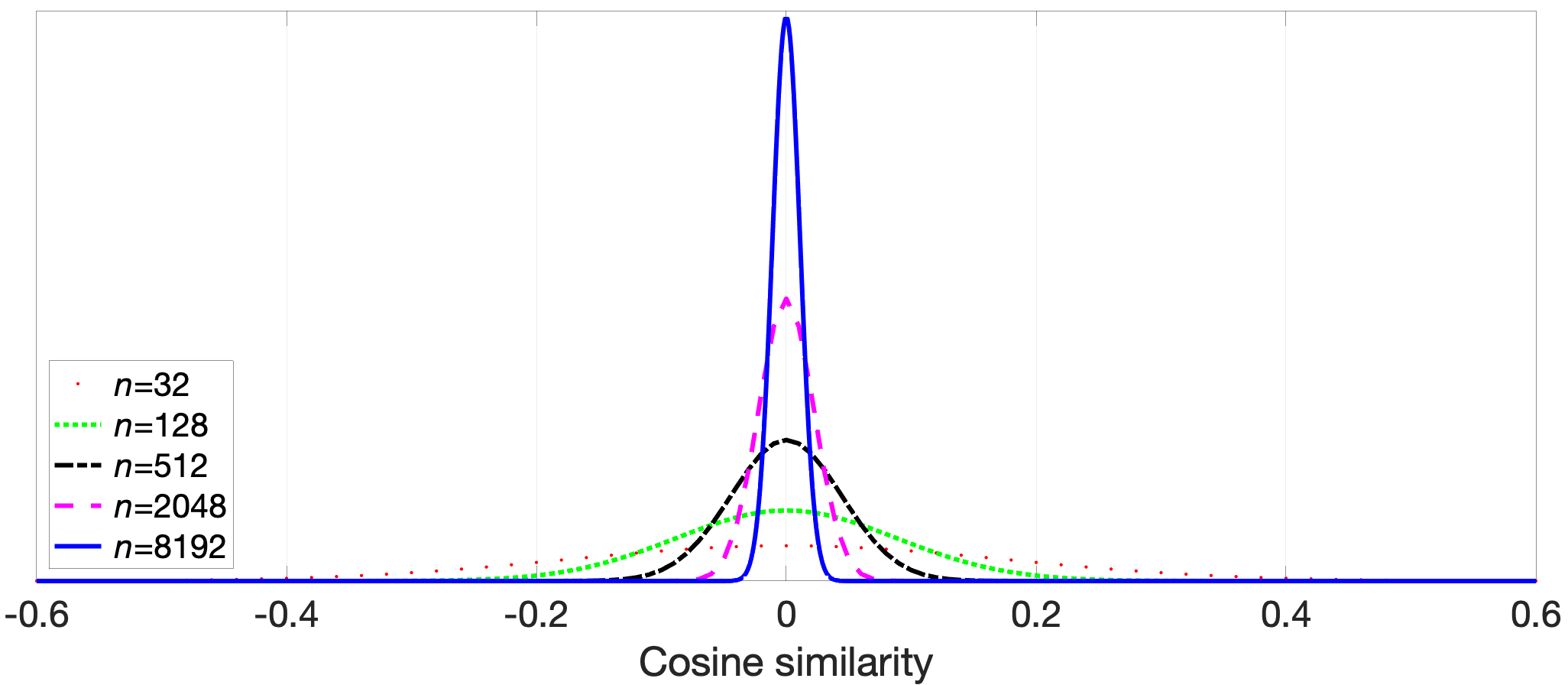}
\caption{
Concentration of measure phenomenon between bipolar random HVs. Lines correspond to probability density functions of cosine similarities for different dimensionalities.
Normal distributions were fitted to pair-wise similarities for $2000$ random HVs.   
}
\label{fig:dimensions}
\end{figure}

As mentioned above, in HDC/VSA random independent HVs generated from some distribution are used for representing objects that are considered independent and dissimilar. 
For example, randomly generated binary HVs with components from the set $\{0,1\}$, with the probability of a 1-component being $p_1 = 1/2$  for dense binary representations~\cite{KanervaHyperdimensional2009} or for sparse binary representations~\cite{RachkovskijStructures2001} with $p_1 = 1/100$ or with a fixed number of $M<<D$ randomly activated components.
Such random HVs are analogous to ``symbols'' in symbolic representations. 
However, the introduction of new symbols in symbolic representations or nodes in localist representations requires changing the dimensionality of representations, whereas in HDC/VSA they are simply introduced as new HVs of fixed dimensionality. So HDC/VSA can accommodate $N$ symbols with $D$-dimensional HVs such that $N>>D$.
This happens because in high-dimensional spaces randomly chosen HVs are quasi-orthogonal to each other. 
The number of exactly orthogonal dimensions in a vector space is equal to its dimensionality $D$, but the number of quasi-orthogonal directions is exponential to the dimensionality.
Exact orthogonality means that the angle between vectors is exactly $90$ degrees, whereas quasi-orthogonality means that the angle between vectors is within some small interval around $90$ degrees. As the dimensionality increases, the number of such quasi-orthogonal vectors increases exponentially even for a very tiny interval around $90$ degrees. So, as the dimensionality increases, the number of quasi-orthogonal vectors increases rapidly while the mathematical properties of the quasi-orthogonal vectors become closer to those of exactly orthogonal vectors.
This mathematical phenomenon is known as concentration of measure~\cite{LedouxConcentration2001}.
The peculiar property of this phenomenon is that quasi-orthogonality converges to exact orthogonality with increased dimensionality of HVs. 
This is sometimes referred to as the ``blessing of dimensionality''~\cite{Gorban2018Blessing}. 
Fig.~\ref{fig:dimensions} provides a visual illustration of the case of bipolar HVs.

If the similarity between objects is important, then HVs are generated in such a way that their similarity characterizes the desired similarity between objects.
The similarity between HVs is measured by standard  vector similarity (or distance) measures.

\subsubsection{Similarity measures}\mbox{}\\
\label{sec:vsa:similarity}
\paragraph{Similarity measure for dense representations}
For dense HVs with real or integer components the most commonly used measures are the Euclidean distance:
\noindent
\begin{equation}
\label{eq:dist:Euc}
\text{dist}_{\text{Euc}} (\mathbf{a}, \mathbf{b}) = \sqrt{ \sum_i (\mathbf{a}_i - \mathbf{b}_i) ^2 } = ||\mathbf{a}-\mathbf{b}||_2;
\end{equation}
\noindent
the dot (inner, scalar) product:
\noindent
\begin{equation}
\label{eq:sim:dot}
\text{sim}_{\text{dot}} (\mathbf{a}, \mathbf{b}) = \sum_i \mathbf{a}_i \mathbf{b}_i = \mathbf{a}^{\top } \mathbf{b};
\end{equation}
\noindent
 and the cosine similarity:
\noindent
\begin{equation}
\label{eq:sim:cos}
\text{sim}_{\text{cos}} (\mathbf{a}, \mathbf{b}) = \frac{\mathbf{a}^{\top } \mathbf{b}}{||\mathbf{a}||_2 ||\mathbf{b}||_2}.
\end{equation}
\noindent
Note that these measures are closely related. 
In fact, when vectors $\mathbf{a}$ and $\mathbf{b}$ are normalized to have unit norm, there is exact correspondence: $\text{sim}_{\text{dot}} (\mathbf{a}, \mathbf{b}) = \text{sim}_{\text{cos}} (\mathbf{a}, \mathbf{b}) = 1- \text{dist}_{\text{Euc}}(\mathbf{a}, \mathbf{b})^2/2$
(as follows from $||\mathbf{a}-\mathbf{b}||_2^2 = ||\mathbf{a}||_2^2 + ||\mathbf{b}||_2^2 - 2\mathbf{a}^{\top } \mathbf{b}$).   
For dense HVs with binary components, the normalized Hamming distance is the most common choice: 
\noindent
\begin{equation}
\label{eq:dist:Ham}
\text{dist}_{\text{Ham}} (\mathbf{a}, \mathbf{b}) = \frac{1}{D} \text{dist}_{\text{Euc}}(\mathbf{a}, \mathbf{b})^2 = \frac{1}{D} | \mathbf{a} \oplus \mathbf{b} |,
\end{equation}
\noindent
where $\oplus$  denotes a component-wise XOR operation. 
If the binary values of $\mathbf{a}$ and $\mathbf{b}$ are mapped to bipolar values by replacing  0-components to $-1$s, the Hamming distance is connected to the dot product as
$\text{sim}_{\text{dot}} (\mathbf{2a-1}, \mathbf{2b-1}) = D(1-2\text{dist}_{\text{Ham}} (\mathbf{a}, \mathbf{b}))$.

\paragraph{Similarity measures for sparse representations}
$\text{sim}_{\text{dot}}$ and $\text{sim}_{\text{cos}}$ are frequently used as similarity measures in the case of sparse HVs, i.e., when the number of nonzero components in HV is small.

Sometimes, Jaccard similarity is also used for this purpose:
\noindent
\begin{equation}
\label{eq:dist:Jac}
\text{sim}_{\text{Jac}} (\mathbf{a}, \mathbf{b}) = \frac{| \mathbf{a} \land \mathbf{b}| }{ |\mathbf{a} \lor \mathbf{b}| },
\end{equation}
\noindent
where $\land$ and $\lor$  denote component-wise AND and OR operations, respectively. 

\subsubsection{Operations}
\label{sec:vsa:operations}
Superposition and binding operations that do not change the HV dimensionality are of particular interest because they allow one to apply the same operations to their resultant HVs. But note that some applications might require changing the dimensionality of representations.
Obviously, in biological systems one would be very surprised if dimensionalities were exactly equal. Assuming that the dimensionality is unchanged is a mathematical convenience for, e.g., using component-wise vector operations -- rather than a requirement.

\paragraph{Superposition}
\label{sec:vsa:operations:superposition}
The superposition is the most basic operation used to form an HV that represents several HVs. 
In analogy to simultaneous activation of neural patterns represented by HVs, this is usually modeled as a disjunction of binary HVs or addition of real-valued HVs. 
See more examples of particular implementations in Section~\ref{sec:vsa:frameworks}.
In HDC/VSA, this operation is known under several names: bundling, superposition, and addition.
Due to its intuitive meaning, below, we will use the term superposition.
The simplest unconstrained superposition (denoted as ${\bf s}$) is simply: 
\noindent
\begin{equation}
\label{eq:super:uncon}
{\bf s}=\mathbf{a} + \mathbf{b} + \mathbf{c}.
\end{equation}
\noindent
In order to be used at later stages, the resultant HV is often required  to preserve certain properties of atomic HVs. 
Therefore some kind of normalization is often used to preserve, for instance, the norm, e.g., Euclidean:
\noindent
\begin{equation}
\label{eq:super:norm}
{\bf s}=\frac{\mathbf{a} + \mathbf{b} + \mathbf{c}}{ || \mathbf{a} + \mathbf{b} + \mathbf{c} ||_2 };
\end{equation}
\noindent
or the type of components as in atomic HVs, e.g., integers in a limited range:
\noindent
\begin{equation}
\label{eq:super:norm:clip}
{\bf s}=f( \mathbf{a} + \mathbf{b} + \mathbf{c}),
\end{equation}
\noindent
where $f()$ is a function limiting the range of values of ${\bf s}$. 
For example, in the case of dense binary/bipolar HVs, the binarization of components via majority rule/sign function is used.
In the case of sparse binary HVs, their superposition increases the density of the resultant HV.
There are operations to ``thin'' the resultant HV and preserve the sparsity (see Section~\ref{sec:SBDR}).
Below, we will also use brackets $\langle \cdot \rangle$ when some sort of normalization is applied.

After superposition, the resultant HV is similar to its input HVs.
Moreover, the superposition of HVs remains similar to each individual HV but the similarity decreases as more HVs are superimposed together. This could be seen as an unstructured and additive kind of similarity preservation.
A fundamental property of the superposition is that it is associative, meaning that the sum is independent of the order
of its components. 
However,~\cite{RiemannNovelHDComputingAlgebra2022} proposed a non-associative implementation of the superposition operation,  which allows building a filter in the temporal (sequential) as well as in the elements’ domains.
Note that the properties above are desiderata for the superposition and, thus, any operation that has these properties shall be seen as an implementation of the superposition operation.

\paragraph{Binding}
\label{sec:binding}
As mentioned above, the superposition operation alone leads to the superposition catastrophe (due to its associative property), i.e., during the recursive application of the superposition, the information about combinations of the initial objects (e.g., grouping) is lost
since, e.g., $(\mathbf{a} + \mathbf{b}) + (\mathbf{c} + \mathbf{d}) = (\mathbf{a} + \mathbf{c}) + (\mathbf{b} + \mathbf{d}) = \dots = \mathbf{a} + \mathbf{b}+ \mathbf{c} + \mathbf{d}$.

This issue  needs to be addressed in order to represent compositional structures (see Section~\ref{sec:motivation:binding}).
A binding operation (denoted as $\circ$ when no particular implementation is specified) is used in HDC/VSA as a solution. 
It is convenient to consider two types of binding: via multiplication-like operation and via permutation operation.
These two types are selected because they are utilized by most applications.
Let us first briefly discuss them before we consider 
concrete implementations of the binding operation by different HDC/VSA models in Section~\ref{sec:vsa:frameworks}. 

{\textbf{Multiplicative binding.}} Multiplication-like operations are used in different HDC/VSA models for implementing the binding operation when two or more HVs should be bound together.
Concrete implementations of this type of binding are rather diverse and depend on a particular HDC/VSA model.  
Examples of multiplication-like binding operations include: conjunction for sparse binary HVs, component-wise multiplication for real-valued HVs, outer product for tensor product representations, and even matrix-vector multiplication when a role is represented by a matrix while a filler is represented by an HV.   
Please refer to Section~\ref{sec:vsa:frameworks} for concrete implementations for a particular HDC/VSA model.
Fig.~1 in~\cite{SchlegelVSAComparison2020} also presents the taxonomy of the most commonly used multiplicative bindings.

The HV obtained after binding together several input HVs depends on all of them. 
Similar input HVs produce similar bound HVs. However, the way the similarity is preserved after the binding is (generally) different from that of the superposition operation~\cite{PlateAnalogy2000, RachkovskijStructures2001}.
We discuss this aspect in detail 
a few paragraphs further down.

In many situations, there is a need for an operation to reverse the result of the binding operation. This operation is called unbinding or release (denoted as $\oslash$).
It allows the recovery of a bound HV~\cite{WiddowsContinuous2015, GosmannVDTB2019}, e.g.,:
\noindent
\begin{equation}
\mathbf{a}= \mathbf{b} \oslash (\mathbf{a} \circ \mathbf{b}).
\end{equation}
\noindent
Here, the binding of only two HVs is used for demonstration purposes. In general, many HVs can be bound to each other. 
As mentioned above, even matrix-vector multiplication can be used to implement binding.
Note that when the unbinding operation is applied to a superposition of HVs, the result will not be exactly equal to the original bound HV. It will also contain crosstalk noise; therefore, the unbinding operation is commonly followed by a clean-up procedure (see Section~\ref{sec:recovery}) to obtain the original bound HV. 
In some models, the binding operation is self-inverse, which means that the unbinding operation is the same as the binding operation. 
We specify how the unbinding operations are realized in different models in Section~\ref{sec:vsa:frameworks}.

{\textbf{Binding by permutation.}} \label{sec:binding:permutation} Another type of HV binding is by permutation, which might correspond to, e.g., some position or some other role of an object. 
We denote the application of a permutation to an HV as 
$\rho(\mathbf{a})$;\footnote{
It is worth mentioning that $\rho(\cdot)$ denotes some arbitrary constant permutation. There may be multiple different permutations used in some applications, in which case they would need to be distinguished by, e.g., different symbols or sub-scripting.
However, it is very common for only one permutation to be used, so the need to distinguish different permutations does not arise.
} if the same permutation is applied $i$ times as $\rho^i(\mathbf{a})$ and the corresponding inverse is then $\rho^{-i}(\mathbf{a})$. 
Note that the permuted HV is  dissimilar to the original one.
However, the use of partial permutations~\cite{KussulPermutative2003, CohenEARP2018} allows preserving some similarity (see Section~\ref{sec:2D:permute}). 
For permutation binding, the unbinding is achieved by inverse permutation. 

The permutation can be implemented 
via a ``permutation vector'' that associates each component's index with its index after permutation or by a matrix multiplication with the corresponding permutation matrix having a single 1-component in each row and column. 
Implementation of the permutation via the multiplication by the permutation matrix highlights some connection to the multiplicative binding implemented by matrix-vector multiplication as in~\cite{GallantRepresenting2013}. 
Though the structure of permutation matrices is very different from the dense random matrices used in~\cite{GallantRepresenting2013} (see also Section~\ref{sec:framework:MBAT}).
However, a typical implementation of the multiplicative binding will take two HVs as an input, while the permutation operation takes as input an HV as well as permutation's identity.
This difference in the type of input arguments explains why sometimes permutations are treated  as the third type of basic HDC/VSA operations (in addition to superposition and binding).

Often in practice a special case of permutation -- cyclic shift -- is used as it is very simple to implement. 
The use of cyclic (as well as non-cyclic) shift for positional binding in HDC/VSA was originally proposed in~\cite{kussul1993information} for  representing 2D structures (images) and 1D structures (sequences).  Permutations were also used for representing the order in~\cite{RachkovskijStructures2001}.
In~\cite{GaylerMAP1998}, the permutation operation was introduced
as essential for representing compositional structures when the binding operation is self-inverse.
In a similar spirit, a permutation was used in~\cite{PlateHolographic1995, JonesMeaning2007} to avoid the commutativity of a multiplicative binding by circular convolution.
In general, a permutation can be used to do so for all commutative operations in all HDC/VSA models (see, e.g., Section 3.6.7 in~\cite{PlateNested1994}).

Later~\cite{SahlgrenOrder2008, KanervaHyperdimensional2009} introduced a primitive for representing a sequence in a compositional HV using multiple applications of the same fixed permutation to represent a position in the sequence.
This primitive was popularized via applications to texts~\cite{SahlgrenOrder2008, JoshiNgrams2016}.

\paragraph{Similarity preservation by operations}
\label{sec:vsa:operations:simil}
For the superposition operation, the ``contribution'' of any input HV on the resultant HV is the same independently of other input HVs. 
In binding, the ``contribution'' of any input HV on the result depends on all other input HVs.

Also, the superposition and binding operations preserve similarity in a different manner. 
There is \textit{\textbf{unstructured similarity}}, which is the similarity of the  
resultant HV to the HVs used as input to the operation. 
The superposition operation preserves the similarity of the resultant HV to each of the superimposed HVs, that is, it preserves the unstructured similarity. 
For example, $\mathbf{d} = \mathbf{a} + \mathbf{b}$ is similar to $\mathbf{a}$ and $\mathbf{b}$. 
In fact, given some assumptions on the nature of the HVs being superimposed, it is possible to analytically analyze the information capacity of HVs (see Section~\ref{sec:capacity} for details).  

Most realizations of the binding operation do not preserve unstructured similarity.
They do, however, preserve \textit{\textbf{structured similarity}}, that is, the similarity of the bound HVs to each other. 
Let us consider two i.i.d. random HVs: $\mathbf{a}$ and $\mathbf{b}$; 
$\mathbf{d} = \mathbf{a} \circ \mathbf{b}$ is similar to 
$\mathbf{d}' = \mathbf{a}' \circ \mathbf{b}'$ if $\mathbf{a}$ is similar to $ \mathbf{a}'$ and $\mathbf{b}$ is similar to $\mathbf{b}'$. 
Thus, most realizations of the binding operation preserve structured similarity in a multiplicative fashion. 
When $\mathbf{a}$ and $\mathbf{b}$ are independent, as well as $\mathbf{a}'$ and $\mathbf{b}'$, then the similarity of $\mathbf{d}$ to $\mathbf{d}'$ is equal to the product of the similarities of $\mathbf{a}$ to $\mathbf{a}'$ and $\mathbf{b}$ to $\mathbf{b}'$. 
For instance, if $\mathbf{a}=\mathbf{a}'$, the similarity of $\mathbf{d}$ to $\mathbf{d}'$ will be equal to the similarity of $\mathbf{b}$ to $\mathbf{b}'$. 
If $\mathbf{a}$ is not similar to $\mathbf{a}'$, $\mathbf{d}$ will have no similarity with $\mathbf{d}'$ irrespective of the similarity between $\mathbf{b}$ and $\mathbf{b}'$ due to the multiplicative fashion of similarity preservation. 
The structured similarity is also preserved when using permutations since $\rho(\mathbf{a})$ is similar to $\rho(\mathbf{a}')$ if $\mathbf{a}$ is similar to $ \mathbf{a}'$.
This type of similarity is different from the unstructured similarity of the superimposed HVs: $\mathbf{d} = \mathbf{a} + \mathbf{b}$ will still be similar to $\mathbf{e} = \mathbf{a} + \mathbf{c}$ even if $\mathbf{b}$ is dissimilar to $\mathbf{c}$.
Finally, it is worth noting that, in practice, it might be necessary to preserve both structured and unstructured similarities (e.g., for tasks involving a similarity search) when binding two objects. In that case, the HVs should be formed in such a way that both similarity types are taken into account.
A simple example of such a representation is $\mathbf{a} + \mathbf{b} + \mathbf{a} \circ \mathbf{b}$ that still preserves some similarity to, e.g., $\mathbf{a} + \mathbf{c} + \mathbf{a} \circ \mathbf{c}$. 

\subsubsection{Representations of compositional structures}
\label{sec:compositional}

As introduced in Section~\ref{sec:connectionist:representations}, compositional structures are formed from objects where the objects can be either atomic or compositional. Atomic objects are the most basic (irreducible) elements of a compositional structure. More complex compositional objects are composed from atomic ones as well as from simpler compositional objects. Such a construction may be considered as a part-whole hierarchy, where lower-level parts are recursively composed to form higher-level wholes (see, e.g., an example of a graph in Section~\ref{sect:graphs:labelled}).

In HDC/VSA, compositional structures are transformed into their HVs using HVs of their elements and the superposition and binding operations introduced above. As mentioned in~\ref{sec:motivation:binding}, it is a common requirement that similar compositional structures be represented by similar HVs. 
Another requirement that was studied intensively in the early days of HDC/VSA (see Section~\ref{PartII-sec:transformation} of Part~II of the survey~\cite{KleykoSurveyVSA2021Part2} and, e.g.,~\cite{NeumannTransformation2002, PlateNested1994, SmolenskyTensor1990}) is the ability to transform representations of compositional structures without requiring the recovery of the structures (as would be the case for the symbolic approach). One possibility
to recover the original representation from its compositional HV, which we describe in the next section, might be an additional requirement. In Section~\ref{sec:data:to:HV} below, we will review a number of approaches to the formation of atomic and compositional HVs.

\subsubsection{Recovery and clean-up}
\label{sec:recovery}

Given a compositional HV, it is often desirable to find the particular input HVs from which it was constructed. We will refer to the procedure implementing this as recovery. This procedure is also known as decoding, reconstruction, restoration, decomposition, parsing, and retrieval.

For recovery, it is necessary to know the set of the input HVs from (some of) which the HV was formed. This set is usually called the dictionary, codebook, clean-up memory or item memory\footnote{
In the context of HDC/VSA, the term clean-up was introduced in~\cite{PlateAlgebra1991} while the term item memory was proposed in~\cite{KanervaAnalogy1998}. }. 
In its simplest form, the item memory is just a matrix storing HVs explicitly, but it can be implemented as, e.g., a content-addressable associative memory~\cite{StewartCleanup2011}.
Note that the stored HVs do not have to be atomic, they could also be compositional representations. 
For example, recent work~\cite{SteinbergStructuredKnowledge2022} studied how to store compositional HVs in content-addressable associative memories \`{a} la Hopfield network. 
Moreover, recent works suggested~\cite{SchmuckHardwareOptimizations2019,KleykoCA2020, EggimannConfigurableHD2021} that it is not always necessary to store the item memory explicitly, as it can be easily rematerialized.
Also, the recovery requires knowledge about the structure of the compositional HV, that is, information about the operations used to form a given HV (see an example below). 

Most of the HDC/VSA models produce compositional HVs not similar to the input HVs due to the properties of their binding operations. So, to recover the HVs involved in the compositional HV, the use of the unbinding operation (Section~\ref{sec:binding} above) will typically be required. If the superposition operation was used when forming the compositional HV, after unbinding the obtained HVs will be noisy versions of the input HVs to be recovered (noiseless version can be obtained in the case when only binding operations were used to form a compositional HV).
Therefore, a ``clean-up'' procedure is used as part of the recovery.  
In its most general form, the clean-up procedure can be viewed as projecting a query HV onto the subspace spanned by the item memory and returning, e.g., the weighted superposition (see examples in~\cite{KentResonatorNetworks2020, FradyResonator2020}).
In practice, however, the most common way of using the clean-up procedure is by finding the HV in the item memory that is most similar to the noisy query HV. 
The item memory performs the nearest neighbor search (i.e., returns an HV corresponding to the winner-take-all activation) that can be implemented by, e.g., comparing the query HV to each of the item memory's HVs or by some kind of search in an associative memory implementing the item memory. 
For HVs representing limited size compositional structures, there are guarantees for the exact recovery~\cite{ThomasHDFoundations2020}.

Let us consider a simple example. The compositional HV was obtained as follows: 
$\mathbf{s} = \mathbf{a} \circ \mathbf{b} +  \mathbf{c} \circ \mathbf{d}$.
For recovery from $\mathbf{s}$, we know that it was formed as the superposition of pair-wise bindings. In the simplest setup, we also know that the input HVs include $\mathbf{a}$, $\mathbf{b}$, $\mathbf{c}$, $\mathbf{d}$. The task is to find which HVs were in those pairs. 
Then, to find out which HV was bound with, e.g., $\mathbf{a}$ we unbind it with $\mathbf{s}$
as 
$ \mathbf{a} \oslash \mathbf{s} $ resulting in $\mathbf{b} + \mathbf{a} \oslash (\mathbf{c} \circ \mathbf{d}) = \mathbf{b} + \mathrm{noise} \approx \mathbf{b}$. 
Then we use the clean-up procedure that returns $\mathbf{b}$ as the closest match (using the corresponding similarity measure).
That way we know that $\mathbf{a}$ was bound with $\mathbf{b}$. In this setup, we immediately know that $\mathbf{c}$ was bound with $\mathbf{d}$. We can check this in the same manner, by first calculating, e.g., $\mathbf{d} \oslash \mathbf{s} $, and then cleaning it up. 
Note that $\mathrm{noise}$ is defined as whatever is quasi-orthogonal to the signals of interest, and this definition is operationalized by populating the item memory with the signals of interest. 
In other words, if the noise term in the above example ($\mathbf{a} \oslash (\mathbf{c} \circ \mathbf{d})$) was an HV in the item memory, then the system would not treat it as noise.

In a more complicated setup, it is not known that the input HVs were $\mathbf{a}$, $\mathbf{b}$, $\mathbf{c}$, $\mathbf{d}$, but the HVs in the item memory are known. So, we take those HVs, one by one, and repeat the operations above. 
A possible final step in the recovery procedure is to recalculate the compositional HV from the reconstructed HVs. The recalculated compositional HV should match the original compositional HV. 
Note that the recovery procedure becomes much more complex in the case where $m$ HVs are bound instead of just pairs. It is easy to recover one of the HVs used in the binding, if the other $m-1$ HVs are known. They can simply be unbound from the compositional HV. 

When the other HVs used in the binding are not known, the simplest way to find the HVs used for binding is to compute the similarity with all possible bindings of HVs from the item memory. The complexity of this search grows exponentially with $m$.
However, there is a recent work~\cite{KentResonatorNetworks2020, FradyResonator2020} that proposed a mechanism called resonator network to address this problem.

\begin{table}[t!]
\tiny
\renewcommand{\arraystretch}{1.0}
\caption{Summary of HDC/VSA models. Each model has its own atomic HVs and operations on them for binding and superposition, and a similarity measure.
}
\centering
\label{ch2:tab:VSAs}
\begin{tabular}{|c|c|c|c|c|c|c|}
\hline
HDC/VSA & \text{Ref.} & \text{Space of atomic HVs} & \text{Binding} & \text{Unbinding} & \text{Superposition}  & \text{Similarity} \\ 
\hline
\text{TPR} & \cite{SmolenskyTensor1990}  & unit HVs   & outer product & tensor-vector inner product  & \makecell{component-wise \\ addition}  & $\text{sim}_{\text{dot}}$ \\ \hline
\text{HRR} & \cite{PlateHolographic1995}  & \text{unit HVs}   & \text{circular convolution} & \makecell{circular correlation} & \makecell{component-wise \\ addition}  & $\text{sim}_{\text{dot}}$ \\ \hline
\text{SMR} & \text{\cite{KellyStructure2013}}  & \makecell{square matrices \\ with unit norm}   &  \makecell{matrix \\ multiplication} &   \makecell{matrix  multiplication \\ with matrix transpose}&  \makecell{component-wise \\ addition}  & $\text{sim}_{\text{cos}}$  \\ \hline
\text{FHRR} & \text{\cite{PlateHolographic2003}}  & \text{complex unitary HVs}   &  \makecell{component-wise \\ multiplication} &  \makecell{component-wise multiplication \\ with complex conjugate} &
\makecell{component-wise \\ addition}  & $\text{sim}_{\text{cos}}$ \\ \hline
\text{SBDR} & \text{\cite{RachkovskijStructures2001}}  & \text{sparse binary HVs}   & \text{context-dependent thinning} &  \makecell{repeated context-\\dependent thinning} & 
\makecell{component-wise \\ disjunction}  & $\text{sim}_{\text{dot}}$ \\ \hline
\text{BSC} & \text{\cite{KanervaFully1997}}  & \text{dense binary HVs}   & \text{component-wise XOR} & \text{component-wise XOR} & \text{majority rule}  & $\text{dist}_{\text{Ham}}$ \\ \hline
\text{MAP} & \text{\cite{GaylerMAP1998}}  & \text{dense bipolar HVs}   &  \makecell{component-wise \\ multiplication} &   \makecell{component-wise \\ multiplication}&  \makecell{component-wise \\ addition}  & $\text{sim}_{\text{cos}}$  \\ \hline
\text{MCR} & \text{\cite{SnaiderModular2014}}  & \text{dense integer HVs}   &  \makecell{component-wise \\ modular addition}  &  \makecell{ component-wise  \\ modular subtraction} & \makecell{component-wise \\ discretized vector sum}  & \makecell{modified \\ Manhattan} \\ \hline
\text{CGR} & \text{\cite{YuUnderstandingHDC2022}}  & \text{dense integer HVs}   &  \makecell{component-wise \\ modular addition}  &  \makecell{ component-wise  \\ modular subtraction} & \makecell{component-wise \\ discretized vector sum}  & \makecell{$\text{sim}_{\text{cos}}$} \\ \hline
\text{MBAT} & \text{\cite{GallantRepresenting2013}}  & \text{dense bipolar HVs}   &  \makecell{vector-matrix \\ multiplication}  &  \makecell{  multiplication with \\ inverse matrix} & \makecell{component-wise \\ addition}  & $\text{sim}_{\text{dot}}$ \\ \hline
\text{SBC} & \cite{LaihoSparse2015}  & \text{sparse binary HVs}  & \makecell{block-wise \\ circular convolution}  & \makecell{block-wise circular convolution \\ with approximate inverse}  & \makecell{component-wise \\ addition}  & $\text{sim}_{\text{dot}}$ \\ \hline
\text{GAHRR} & \text{\cite{AertsGeometric2009}}  & \text{unit HVs}   & \text{geometric product}  & \makecell{geometric product \\ with inverse} & \makecell{component-wise \\ addition}  & \makecell{unitary \\ product} \\ \hline
\end{tabular}
\end{table}

\subsection{The HDC/VSA models}
\label{sec:vsa:frameworks}

In this section, various 
HDC/VSA models are overviewed. For each model we provide a format of employed HVs  and the implementation of the basic operations introduced in Section~\ref{sec:vsa:operations} above. Table~\ref{ch2:tab:VSAs} provides the summary of the  models (see also Table~1 in~\cite{SchlegelVSAComparison2020})\footnote{Note that the binding via the permutation operation is not specified in the table. 
That is because it can be used in any of the models even if it was not originally proposed to be a part of it. 
Here, we limit ourselves to specifying only the details of different models but see, e.g.,~\cite{SchlegelVSAComparison2020} for some comparisons of seven different models from Table~\ref{ch2:tab:VSAs} (the work did not cover the GAHRR, MCR, and TPR models) and some of their variations ($2$ for HRR, $3$ for MAP, and $2$ for SBDR).
Note also that in the table we specify the similarity measure used in the source references. 
However, as indicated in Section~\ref{sec:vsa:similarity}, there is a number of alternatives. 
Thus, the mentioned similarity measure by no means should be considered as a prescription. Instead, in practice, any similarity measure that can be applied to the particular model can be used.
}.  

\subsubsection{A guide on the navigation through the HDC/VSA models}
Before exposing a reader to the details of each HDC/VSA model, it is important make a comment on the nature of their diversity and enable an intuition behind selecting the best model for the practical usage.  

The current diversity of the HDC/VSA models is a result of the evolutionary development of the main vector symbolic paradigm by independent research groups and individual researchers. Initially, the diversity comes from different initial assumptions, variations in the neurobiological inspiration and the particular mathematical background of the originators.  
Therefore,  from a historical perspective, the question of selecting a candidate for the best model is ill-posed. Recent work~\cite{SchlegelVSAComparison2020} started to perform a systematic experimental comparison between models. We emphasize the importance of further investigations in this direction in order to facilitate conscious choice of one or another model for a given problem and to raise HDC/VSA to the level of matured engineering discipline.

However, the usage of different models could already be prioritized at this moment using the target computing hardware perspective. The recent developments of unconventional computing hardware aim at improving the energy efficiency over the conventional von Neumann architecture in AI applications. 
Another driving factor is reliability. 
As conventional hardware for the von Neumann architecture gets smaller, it eventually becomes unreliable, so there is a need for computing frameworks that are robust to unreliable hardware and HDC/VSA is a promising candidate. 
The availability of such a robust computing framework also opens the door to other types of hardware implementation that are inherently unreliable and would not be
suitable for a von Neumann architecture.
Various unconventional hardware platforms (see Section~\ref{disc:develop:implmentation})
deliver great promises in moving the borders of energy efficiency and operational speed beyond the current standards.

Independently on the type of the computing hardware, any HDC/VSA model can be seen as an abstraction of the algorithmic layer and can thus be used for designing computational primitives, which can then be mapped to various hardware platforms using different models~\cite{KleykoComputingParadigm2021}.  In the next subsections, we present the details of the currently known HDC/VSA models.

\subsubsection{Tensor Product Representations}
\label{sec:framework:TPR}

The Tensor Product Representations model (in short, the TPR model or just TPR) is one of the earliest models within HDC/VSA. 
The TPR model was originally proposed by Smolensky in~\cite{SmolenskyTensor1990}. 
However, it is worth noting that similar ideas were also presented in~\cite{MizrajiContext1989, MizrajiCalculus1992} around the same time but  received much less attention from the research community.
Atomic HVs are vectors selected uniformly at random from the Euclidean unit sphere $S^{(D-1)}$. 
The binding operation is an outer product (a generalized outer product) of HVs. 
So, the dimensionality of bound HVs grows exponentially with their number (e.g., 2 bound HVs have the dimension $D^2$, 3 bound vectors have the dimension $D^3$, and so on). 
The vectors need not be of the same dimension. 
This points to another important note that the TPR model may or may not be categorized as an HDC/VSA model depending on whether the fixed dimensionality of representations is considered a compulsory attribute of an HDC/VSA model or not.
Historically, HDC/VSA models have required the operator result HVs to be the same dimensionality as the argument HVs. This was for engineering feasibility and mathematical convenience. 
It is useful to think of HDC/VSA as TPR with the result projected into a lower dimensional space in a way that retains some useful properties of TPR. One way is to think of TPR as a predecessor of HDC/VSA.

The superposition is implemented by (tensor) addition. 
Since the dimensionality grows, a recursive application of the binding operation is challenging. 
The resultant tensor also depends on the order in which the HVs are presented. 
Binding  similar HVs will result in similar tensors. 
The unbinding is realized by taking the tensor product representation of a compositional structure and extracting from it the HV of interest.
For linearly independent HVs, the exact unbinding is done as the tensor multiplication by the unbinding HV(s). 
Unbinding HVs are obtained as the rows of inverse of the matrix with atomic HVs in columns.
Approximate unbinding is done using an atomic HV instead of the unbinding HV. 
If the atomic HVs are orthonormal, this results in the exact unbinding.

Though the similarity measure was not specified, we assume it to be $\text{sim}_{\text{cos}} /\text{dist}_{\text{Euc}}$.

\subsubsection{Holographic Reduced Representations}
\label{sec:vsa:hrr}

The Holographic Reduced Representations model (HRR) was developed by Plate in the early 1990s~\cite{PlateAlgebra1991, PlateNested1994, PlateHolographic1995}.

The HRR model was inspired by Smolensky's TPR model~\cite{SmolenskyTensor1990}, Hinton's ``reduced descriptions''~\cite{Hinton1990mapping},  and many other works, e.g.,~\cite{LonguetHolographic1968, BorsellinoConvolution1973,metcalfe_eich_composite_1982,murdock_theory_1982,schonemann_algebraic_1987} to mention a few.
Note that due to the usage of HRR in Semantic Pointer Architecture Unified Network~\cite{EliasmithBuildBrain2013}, sometimes the HRR model is also referred to as the Semantic Pointer Architecture (SPA). The most detailed source of information on HRR is Plate's book~\cite{PlateHolographic2003}.

In HRR, the atomic HVs  for representing dissimilar  objects are real-valued and their components are independently generated from the normal distribution with mean $0$ and variance $1/D$.
For large $D$, the Euclidean norm is close to $1$.\footnote{
It might be practical to use more constraints when forming atomic HVs, such as using HVs whose entries of fast Fourier transform have unit magnitude (see, e.g.,~\cite{KomerSpatial2020, GanesanLearning2021}). 
} 
The binding operation is defined on two HVs ($\mathbf{a}$ and $\mathbf{b}$) and implemented via the circular convolution, which projects the outer product back onto $D$-dimensional space: 
\noindent
\begin{equation}
\mathbf{a} \circ \mathbf{b}  \equiv z_j = \sum_{k=0}^{D-1}  b_k a_{j-k \mod D},
\end{equation}
\noindent
where $z_j$ is the $j$th component of the resultant HV $\mathbf{z}=\mathbf{a} \circ  \mathbf{b}$.

The circular convolution multiplies norms, and it approximately preserves the unit norms of input HVs.
The bound HV is not similar to the input HVs. 
However, the bound HVs of similar input HVs are similar.
The unbinding is done by the circular correlation of the bound HV with one of the input HVs. 
The result is noisy, so a clean-up procedure (see Section~\ref{sec:recovery}) is required.
There is also a recent proposal in~\cite{GosmannVDTB2019} for an alternative realization of the binding operation called Vector-derived Transformation Binding (VTB).
It is claimed to better suit implementations in spiking neurons and was demonstrated to recover HVs of sequences (transformed with the approach from~\cite{RecchiaSemantic2015}) better than the circular convolution.
Another proposal related to HRR is Square Matrix Representations (SMR)~\cite{KellyStructure2013}, which use random square matrices with unit norm and matrix multiplication as the binding operation (see Table~\ref{ch2:tab:VSAs} for summary).  
Finally, in~\cite{HirataniQuadraticBinding2022} a general formulation of ``quadratic binding'' that subsumes, e.g., outer product-based binding and circular convolution-based binding was proposed.
It can, for example, be used adjust the dimensionality of the bound HV to be within $D$ and $D^2$.

The superposition operation is component-wise addition.
Often, the normalization is used after the addition to preserve the unit Euclidean norm of compositional HVs. 
In HRR, both superposition and binding operations are commutative. The similarity measure is $\text{sim}_{\text{dot}}$ or $\text{sim}_{\text{cos}}$.

\subsubsection{Matrix Binding of Additive Terms}
\label{sec:framework:MBAT}

In the Matrix Binding of Additive Terms model (MBAT)~\cite{GallantRepresenting2013} by Gallant and Okaywe, it was proposed to implement the binding operation by matrix-vector multiplication. Such an option was also briefly mentioned in~\cite{PlateHolographic2003}. 
Generally, it can change the dimensionality of the resultant HV, if needed.

Atomic HVs are dense and their components are randomly selected  independently  from $\{-1,+1\}$. 
HVs with real-valued components are also possible. 
The matrices are random, 
e.g., with elements also from $\{-1,+1\}$. Matrix-vector multiplication results can be binarized by thresholding.  

The properties of this binding are similar to those of most other models (HRR, MAP, and BSC). In particular, similar input HVs will result in similar bound HVs. The bound HVs are not similar to the input HVs (which is especially evident here, since even the HV's dimensionality could change). 
The similarity measure is either $\text{dist}_{\text{Euc}}$ or $\text{sim}_{\text{dot}}$. The superposition is a component-wise addition, which can be normalized. The unbinding could be done using the (pseudo) inverse of the role matrix (or the inverse for square matrices, guaranteed for orthogonal matrices, as in \cite{TisseraQAMBAT2014}). 

In order to check if a particular HV is present in the sum of HVs that were bound by matrix multiplication, that HV should be multiplied by the matrix and the similarity of the resultant HV should be calculated and compared to a similarity threshold.

\subsubsection{Fourier Holographic Reduced Representations}
\label{sec:vsa:fhrr}

The Fourier Holographic Reduced Representations model (FHRR)~\cite{PlateNested1994, PlateHolographic2003} was introduced by Plate as a model inspired by HRR where the HVs' components are complex numbers. 
The atomic HVs are complex-valued random vectors where 
each vector component can be considered as a phase angle (phasor) randomly selected independently  from the uniform distribution over $(0, 2\pi]$. 
Usually, the unit magnitude of each component is used (unitary HVs).
The similarity measure is the mean of  sum of cosines of  angle differences. The binding operation is a component-wise complex multiplication, which is often referred to as Hadamard product. 
The unbinding is implemented via the binding with an HV conjugate (component-wise angle subtraction modulo $2\pi$).
The clean-up procedure is used similarly to HRR.

The superposition is a component-wise complex addition, which can be followed by the magnitude normalization such that all components have the unit magnitude.

\subsubsection{Binary Spatter Codes}
\label{sec:vsa:bsc}

The Binary Spatter Codes model (BSC)\footnote{This acronym was not used in the original publications but we use it here to be consistent with the later literature.} was  developed in a series of papers by Kanerva in the mid 1990s~\cite{KanervaSpatter1994, KanervaFamily1995, KanervaOrdered1996, KanervaFully1997}.
As noted in~\cite{PlateCommon1997}, the BSC model could be seen as a special case of the FHRR model where angles are restricted to $0$ and $\pi$. 
In BSC, atomic HVs are dense binary HVs with components from $\{0,1\}$. 
The superposition is a component-wise addition, thresholded (binarized) to obtain  an approximately equal density of ones and zeros in the resultant HV. 
This operation is usually referred to as the majority rule or majority sum. 
It selects zero or one for each component depending on whether the number of zeros or ones in the summands is higher and ties are broken, e.g., at random. In order to have a deterministic implementation of the majority rule, it is common to assign a fixed random HV, which is included in the superposition when the number of summands is even.
Another alternative is to use an exclusive OR operation on the nearby components of the first and last summands to break a tie~\cite{HannaganHolographic2011}. 

The binding is a component-wise exclusive OR (XOR),
which is equivalent to taking the diagonal of the XOR outer product of the arguments. That is, it is another example of projecting the outer product onto $D$ dimensions.
The bound HV is not similar to the input HVs but 
bound HVs of two similar input HVs are similar.
The unbinding operation is also XOR, 
meaning that BSC has a binding operation with a self-inverse binding property, which may be an important property for choosing an appropriate model for a specific application.
The connection to the FHRR model also indicates that only phase angles with binary discrete phase values can support the self-inverse binding property. But will not support fractional exponentiation (see Section~\ref{sec:scalars:vectors:compos}) because they do not have sufficient resolution of the phase angles.
The clean-up procedure is used similar to the models above.
The similarity measure is often $\text{dist}_{\text{Ham}}$.

\subsubsection{Multiply-Add-Permute}

The Multiply-Add-Permute model (MAP) was proposed by Gayler~\cite{GaylerMAP1998}. 
It has several variants (see~\cite{SchlegelVSAComparison2020}); the bipolar one is isomorphic to the BSC model.
In the MAP model, atomic HVs are dense bipolar HVs with components from $\{-1,1\}$. 
This is used more often than the originally proposed version with the real-valued components from $[-1,1]$.

The binding is component-wise multiplication, which is equivalent to taking the diagonal of the outer product of the arguments.
The bound HV is not similar to the input HVs. 
However, bound HVs of two similar input HVs are similar. 
The unbinding is also component-wise multiplication and requires knowledge of the bound HV and one of the input HVs for the case of two bound input HVs. 

The superposition is a component-wise addition.
The result of the superposition can either be  left as is, normalized using the Euclidean norm, or binarized to $\{-1,1\}$ with the sign function where ties for 0-components should be broken randomly (but deterministically for a particular input HV set). The choice of normalization is use-case dependent. 
The similarity measure is either $\text{sim}_{\text{dot}}$ or
$\text{sim}_{\text{cos}}$.

\subsubsection{Sparse Binary Distributed Representations}
\label{sec:SBDR}

The Sparse Binary Distributed Representations model (SBDR)~\cite{RachkovskijWorldModel2013},  also known  as Binary Sparse Distributed Codes~\cite{RachkovskijStructures2001, RachkovskijBinding2001}, proposed by Kussul, Rachkovskij, et al. emerged as a part of Associative-Projective Neural Networks~\cite{KussulAPNN1991} (see Section~\ref{PartII-sec:cognitive:APNN} in Part~II of this survey~\cite{KleykoSurveyVSA2021Part2}).
There are two variants of the binding operation in the SBDR model: Conjunction and Context-Dependent Thinning. 
The idea of both variants is that they produce a resultant HV where the 1-components are a subset of the 1-components of the input HVs. 
This subset depends on all of the input HVs, so that (with the proper control of the sparsity) a particular combination of the input HVs results in a unique resultant HV, thus binding the input HVs and preserving the information on the input HVs used to obtain the resultant HV.

\paragraph{Conjunction-Disjunction}
The Conjunction-Disjunction is one of the earliest HDC/VSA models. It uses binary vectors for atomic HVs~\cite{RachkovskijAudio1990} (the model was also proposed independently in~\cite{KanervaBoolean1998}).
Component-wise conjunction of HVs is used as the binding operation.  Component-wise disjunction of HVs is used as the superposition operation. 
Both basic operations are defined for two or more HVs. 
Both operations produce HVs similar to the HVs from which they were obtained (unstructured similarity). That is, the bound HV is similar to the input HVs, unlike most of the other known binding operations, except for Context-Dependent Thinning (see below).

The HVs resulting from both basic operations are similar if their input HVs are similar (structured similarity). 
The operations are commutative so the resultant HV does not depend on the order in which the input HVs are presented. 
However, as noted in Section~\ref{sec:binding:permutation}, this could easily be changed with the use of some random permutations to represent the order of arguments.

The probability of 1-components in the resultant HV can be controlled by the probability of 1-components in the input HVs. Sparse HVs allow using efficient algorithms such as inverted indexing or auto-associative memories for similarity search. 
Since conjunction reduces the density of the resultant HV compared to the density of the input HVs, it could be considered as another form of ``reduced descriptions'' from ~\cite{Hinton1990mapping}.
At the same time, the recursive application of such binding leads to HVs with all components  set to zero. However, the superposition by disjunction increases the number of 1-components. These operations can therefore be used to compensate each other (up to a certain degree; see the next section). A modified version of the Conjunction-Disjunction model preserves the density of 1-components in the resultant HV as shown in the following subsubsection.

\paragraph{Context-Dependent Thinning}
The Context-Dependent Thinning (CDT) procedure proposed by Rachkovskij and Kussul~\cite{RachkovskijBinding2001} can be considered either as a binding operation or as a combination of superposition and binding. 
It was already used in the 1990s~\cite{RachkovskijAudio1990,KussulSequences1991, RachkovskijTexture1991, KussulAPNN1991} under the name ``normalization'' since it approximately maintains the desired density of 1-components  in the resultant HV. 
This density, however,  also determines the ``degree'' or ``depth'' of the binding as discussed in~\cite{RachkovskijBinding2001}. 

The CDT procedure is implemented as follows.
First, $m$ input binary HVs are superimposed by component-wise disjunction as:
\noindent
\begin{equation}
\mathbf{z}=\vee_{i=1}^m \mathbf{a}_i.  
\end{equation}
\noindent
This leads to an increased number of  1-components in the resultant HV and the input HVs are not bound yet. 
Second, in order to bind the input HVs, the  resultant HV $\mathbf{z}$ from the  first stage is permuted and the permuted HV is conjuncted with the HV from the first stage.
This binds the input HVs and reduces the density of  1-components.  
Third, $T$ such conjunctive bindings can be obtained by using different random permutations or by recursive usage of a single permutation. They are superimposed by disjunction to produce the resultant HV. 
These steps are described by the following equation:
\noindent
\begin{equation}
\langle \mathbf{z} \rangle=\vee_{k=1}^T (\mathbf{z} \wedge \rho_{k}( \mathbf{z}))= \mathbf{z} \wedge (  \vee_{k=1}^T \rho_{k} (\mathbf{z})),
\end{equation}
\noindent
where $\rho_{k}$ denotes the $k$th random permutation.
The density of the resultant HV depends on the number of permutation-disjunctions $T$. 
If many of them are used, we eventually get the HV from the first stage, i.e., with no binding of the input HVs.
The described version of the CDT procedure is called ``additive''.
Several alternative versions of the  CDT procedure were proposed in~\cite{RachkovskijBinding2001}.
The CDT procedure preserves the information about HVs that were bound together 
by their remaining 1-components. 
The CDT procedure is commutative.
It is defined for several input HVs. 
As well as Conjunction-Disjunction, the CDT procedure preserves both unstructured and structured similarity.

Due to the fact that both variants of binding in SBDR preserve unstructured similarity, the resultant HV is similar to the input HVs.
This leads to $\langle \mathbf{a}, \mathbf{b}  \rangle$ being similar to $\langle \mathbf{a}, \mathbf{c}  \rangle$, which is not the case for, e.g., $\mathbf{a} \circ \mathbf{b}$ and $\mathbf{a} \circ \mathbf{c}$ in most other implementations of multiplicative binding.  
Here, the preservation of the unstructured similarity allows recovering the input HVs from the resultant HV by using the similarity search in the item memory containing all possible input HVs. 
This does not require any unbinding operation. However, it should be noted that an analog of unbinding could be constructed for SBDR  using repeated binding. This was not investigated in details (except for some preliminary results in~\cite{RachkovskijPhD1990, KussulSequences1991}).

Finally, it is important to take note of the character of the structured similarity preservation by the CDT procedure.  It depends on the thinning depth $T$. 
For a large $T$, the result is close to the superposition of the input HVs, and so the character of the structured similarity is close to additive. The smaller $T$ is, the closer the character of the structured similarity is to the multiplicative one.

\subsubsection{Sparse Block Codes}
The main motivation behind the Sparse Block Codes model (SBC), proposed by Laiho et al.~\cite{LaihoSparse2015, FradySDR2020}, is to use sparse binary HVs as in SBDR but with a binding operation where the bound HV is not similar to the input HVs (in contrast to SBDR).

Similar to SBDR, atomic HVs in the SBC model are sparse and binary with components from $\{0,1\}$. HVs, however, are imposed with an additional structure. An HV is partitioned into blocks of equal size (so that the HV's dimensionality is a multiple of the block size). In each block, there is only one nonzero component, i.e., the activity in each block is maximally sparse as in Potts associative memory, see, e.g.,~\cite{GritsenkoAMSurvey2017}.

In SBC, the binding operation is defined for two HVs and implemented via circular convolution, which is applied block-wise. The unbinding is done by the block-wise circular correlation of the bound HV with one of the input HVs. 
Note that with maximally sparse blocks, the binding operation is equivalent to the modulo sum of the corresponding indices~\cite{LaihoSparse2015}, which demonstrates that each block is effectively a phase angle that is discretized to the block size. 
When the block is not maximally sparse, it functions more like a superposition of phase angles.

The superposition operation is a component-wise addition.
If the compositional HV is going to be used for binding, it might be binarized by leaving only one of the components with the largest magnitude within each block active. If there are several components with the same largest magnitude then a deterministic choice (e.g., the component with the largest position number could be chosen) can be made. The similarity measure is $\text{sim}_{\text{dot}}$ or
$\text{sim}_{\text{cos}}$.

\subsubsection{Modular Composite Representations}

The Modular Composite Representations model (MCR)~\cite{SnaiderModular2014} proposed by Snaider and Franklin shares some similarities with FHRR, BSC, and SBS. Components of atomic HVs are integers drawn uniformly from some limited range (denoted as $r$), which is a parameter of the MCR model. 
Note that this model is also connected to the FHRR model since the components of HVs can be seen as phase angles that are  discretized to $r$ values.

The binding operation is defined as component-wise modular addition (the module value depends on the range limit), which generalizes XOR used for binding in BSC. Binding properties are similar to most of other models. The unbinding operation is the component-wise modular subtraction.

The similarity measure is a variation of the Manhattan distance~\cite{SnaiderModular2014}: 
\noindent
\begin{equation}
\sum_{i=1}^D \min(\mod_r(a_i-b_i),\mod_r(b_i-a_i) ).
\end{equation}
\noindent

The superposition operation resembles that of FHRR. Integers are interpreted as discretized phase angles on a unit circle. First, phase angles are superimposed for each component (i.e., vector addition is performed). Second, the result of the superposition is normalized by setting the magnitude to one and the phase to the nearest phase corresponding to an integer from the defined range.  

Finally, there is also a recent proposal of the Cyclic Group Representations model (CGR)~\cite{YuUnderstandingHDC2022} that highly resembles the idea of the MCR model. 
As it also defines a model that operates with integer-valued HVs but uses slightly different definitions for the similarity measure and the superposition operation.

\subsubsection{Geometric Analogue of Holographic Reduced Representations}

As its name suggests, the Geometric Analogue of Holographic Reduced Representations model (GAHRR)~\cite{AertsGeometric2009} (proposed by Aerts, Czachor and De Moor) was developed as a model alternative to HRR. An earlier proposal was a geometric analogue to BSC~\cite{AertsGABSC2006}. The main idea is to  reformulate HRR in terms of geometric algebra. 
The binding operation is implemented via the geometric product (a generalization of the outer product). 
Because of not projecting the geometric product onto a lower dimensional space, GAHRR is conceptually closest to the TPR model.
The superposition operation is component-wise addition.

So far, GAHRR is mainly an interesting theoretical effort, as its advantages over conceptually simpler models are not particularly clear, but it might become more relevant in the future. 
Readers interested in GAHRR are referred to the original publications. 
An introduction to the model is given in ~\cite{AertsGeometricProduct2008, AertsGeometric2009, PatykGAComputing2010, PatykContinuous2011}, examples of representing data structures with GAHRR are in ~\cite{PatykInformationGAc2011} and some  experiments comparing GAHRR to other HDC/VSA models are presented in ~\cite{PatykPreservingInfo2011, PatykComparison2011}.

\subsection{Information capacity of HVs}
\label{sec:capacity}

An interesting question, which is often brought up by newcomers to the field, is how much information one could store in the superposition of HVs.\footnote{
This question is almost always posed with respect to the superposition. 
It is interesting, however, to consider whether the question makes sense when applied to multiplicative binding. 
It has recently been mentioned in the context of studying resonator networks (see Section~\ref{sec:recovery})~\cite{KentResonatorNetworks2020}.
}
This value is called the information capacity of HVs.
Usually, it is assumed that atomic HVs in superposition are random and dissimilar to each other. 
In other words, one could think about such superposition as an HV representing, e.g., a set of symbols. In general, the capacity depends on parameters such as the number of symbols in the item memory (denoted as $N$), the dimensionality of HVs ($D$), the type of the superposition operation, and the number of HVs being superimposed ($m$). 

Early results on the capacity were given in~\cite{PlateNested1994, PlateHolographic2003}. 
Some ideas for the case of binary/bipolar HVs in BSC, MAP, and MBAT were also presented in~\cite{KleykoHolographic2017, GallantRepresenting2013}. 
The capacity of SBDR was analyzed in~\cite{KleykoTradeoffs2018}.
The most general and comprehensive analysis of the capacity of different HDC/VSA models (and also some classes of recurrent neural networks) was recently presented in~\cite{FradyCapacity2018}. 
The key idea of the capacity theory~\cite{FradyCapacity2018} can be illustrated by the following scenario when an HV to be recovered contains a valid HV from the item memory and crosstalk noise from some other elements of the compositional HV (e.g., from role-filler bindings, see Section~\ref{sec:key:value}). 
Statistically, we can think of the problem of recovering the correct atomic HV from the item memory as as a detection problem with two normal distributions: hit and reject; where hit corresponds to the distribution of  similarity values (e.g., $\text{sim}_{\text{dot}}$) of the correct atomic HV while reject is the distribution of all other atomic HVs (assuming all HVs are random).
Each distribution is characterized by its corresponding mean and standard deviation: $\mu_{h}$ \& $\sigma_{h}$ and $\mu_{r}$ \& $\sigma_{r}$, respectively. 
Given the values of $\mu_{h}$, $\sigma_{h}$, $\mu_{r}$, and $\sigma_{r}$, we can compute the expected accuracy ($p_{corr}$) of retrieving the correct atomic HV according to:
\noindent
\begin{equation}
p_{corr}= \int_{-\infty}^{\infty} \frac{dx}{\sqrt{2 \pi} \sigma_{h}} e^{-\frac{ (x-(\mu_{h}-\mu_{r}))^2 }{2 \sigma_{h}^2}} ( \Phi(x,0, \sigma_{r}))^{N-1},
 \label{eq:pcorr:orig}
 \end{equation}
\noindent
where $\Phi(x)$ is the cumulative Gaussian and $N$ denotes the size of the item memory.  

\begin{figure}[t]
\centering
\includegraphics[width=1.0\columnwidth]{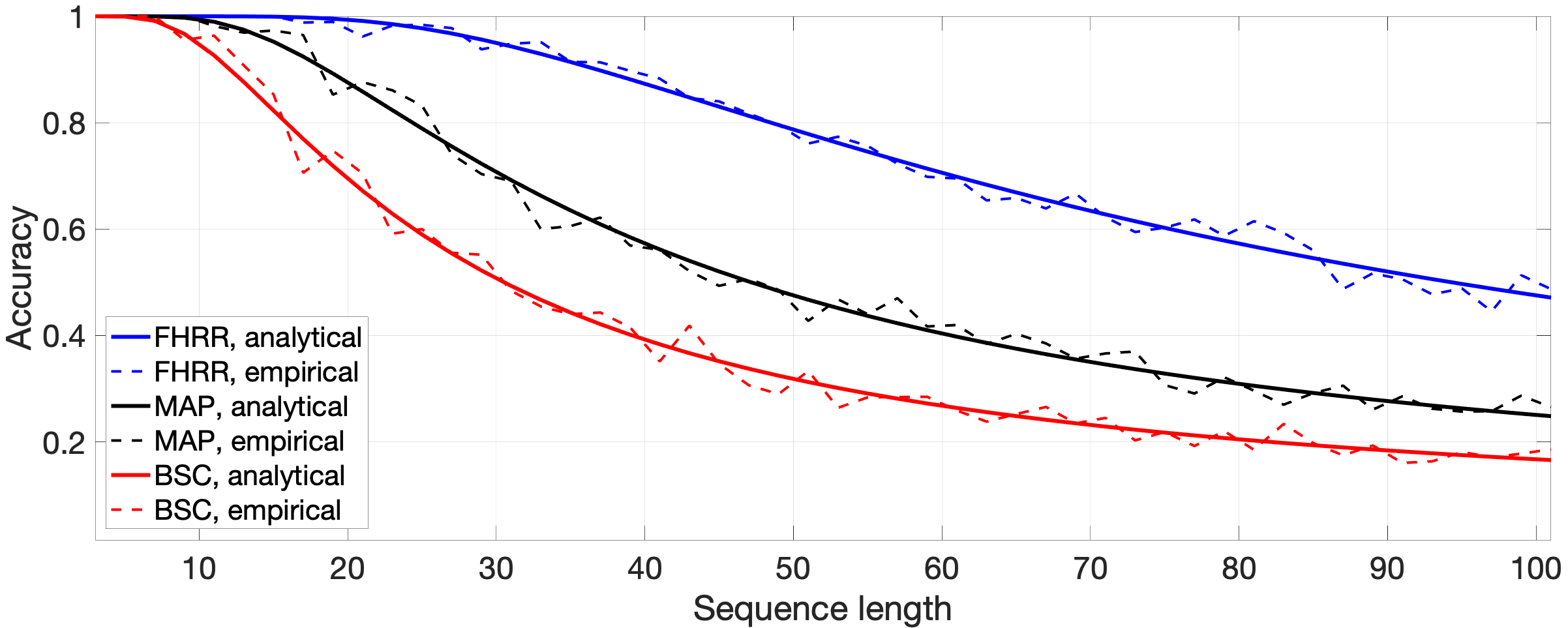}
\caption{
The analytical and empirical accuracies of recovering sequences from HVs against the sequence length for three HDC/VSA models, $D=256$, $N=64$.  
The reported empirical accuracies were averaged over $5$ simulations with randomly initialized item memories. 
}
\label{fig:capacity}
\end{figure}

Fig.~\ref{fig:capacity} depicts the accuracy of retrieving sequence elements from its compositional HV (see Section~\ref{sec:sequences}) for three HDC/VSA models: BSC, MAP, and FHRR.
The accuracies are obtained either empirically or with (\ref{eq:pcorr:orig}). 
As we can see, the capacity theory (\ref{eq:pcorr:orig}) predicts the expected accuracy very accurately.

The capacity theory~\cite{FradyCapacity2018} has recently been extended to also predict the accuracy of HDC/VSA models in classification tasks~\cite{KleykoPerceptron2020}. 
\cite{ThomasHDFoundations2020} presented bounds for the perfect retrieval of sets and sequences from their HVs. Recent works~\cite{MirusCapacity2020, SchlegelVSAComparison2020} have reported empirical studies of the capacity of HVs. Some of these results can be obtained analytically using the capacity theory.  
Additionally,~\cite{Summers2018, FradyCapacity2018, KimHDM2018, HerscheHDMClassifier2021} elaborated on methods for recovering information from compositional HVs beyond the standard nearest neighbor search in the item memory, reaching to the capacity of up to 1.2 bits/component~\cite{HerscheHDMClassifier2021}.
Additional improvements to increase the capacity up to 1.4 bits/component were reported in~\cite{KleykoDecoding2022}.    
This work also provided a taxonomy of existing methods for recovering information from compositional HVs and reported an empirical comparison of these methods.

The works above were focused on the case where a single HV was used to store information but as it was demonstrated in~\cite{DanihelkaAssociative2016} the decoding from HVs can be improved if the redundant storage is used.
As an example, the ``multiset intersection'' circuit in~\cite{GaylerIsomorphism2009} effectively does this by summing over multiple copies of the same data (derived from multiple data permutations) to average away noise.

\section{Data transformation to HVs}
\label{sec:data:to:HV}

As mentioned above, the similarity of atomic symbolic and localist representations is all-or-none: identical symbols are considered maximally similar, whereas different symbols are considered dissimilar. On the other hand, the similarity of structures consisting of symbols (such as sequences, graphs, etc.) is often calculated using computationally expensive procedures such as an edit distance.
HVs are vector representations, and, therefore, rely on simple vector similarity measures that can be calculated component-wise and provide the resultant similarity value immediately.  
This also concerns similarity of compositional structures. Thus, for example, the similarity of relational structures, taking into account the similarity of their elements, their grouping, and order, can be estimated by measuring the similarity of their HVs, without the need for explicit decomposition such as following edges or matching vertices of underlying graphs.
Moreover, HDC/VSA can overcome problems with  symbolic and localist representations concerning the lack of semantic basis (i.e., the lack of immediate similarity of objects; see Section~\ref{sec:motivation:types}).
HDC/VSA overcomes these problems by explicit similarity in their representations, because HVs do not have to represent similarity in all-or-none fashion.
These promises, however, bring the problem of designing concrete transformations that form HVs of various compositional structures such that the similarity between their HVs will reflect the similarity of the underlying objects.

In this section, we consider how data of various types are transformed to HVs to create such representations. 
The data types include symbols (Section~\ref{sec:symb}), sets (Section~\ref{sec:sets}), role-filler bindings (Section~\ref{sec:key:value}), numeric scalars and vectors (Section~\ref{sec:scalars:vectors}), sequences (Section~\ref{sec:sequences}), 2D images (Section~\ref{sec:2Dimages}), and graphs (Section~\ref{sect:graphs}). 

\subsection{Symbols and Sets}
\label{sec:symb:sets}

\subsubsection{Symbols}
\label{sec:symb}

As mentioned in Section~\ref{sect:atom:repr}, the simplest data type is a symbol.
Usually, different symbols are considered dissimilar, therefore, using i.i.d. random HVs for different symbols is a standard way of transforming symbols into HVs. 
Such HVs correspond to symbolic representations (Section~\ref{sec:motivation:types}) since they behave like symbols in the sense that the similarity between the HV and its copy is maximal (i.e., the same symbol) while the similarity between two i.i.d. random HVs is minimal (i.e., different symbols).

\subsubsection{Sets}
\label{sec:sets}

In localist representations, sets are often represented as binary ``characteristic vectors'', where each vector component corresponds to a particular symbol from the universe of all possible symbols. Symbols present in a particular set are represented by one in the corresponding vector component. In the case of multisets, the values of the components of the characteristic vector are the counters of the corresponding symbols. 

In HDC/VSA, a set of symbols is usually represented by an HV formed by the superposition of the symbols' HVs. 
The compositional HV preserves the similarity to the symbols' HVs. Notice that Bloom filter~\cite{Bloom1970space} -- a well-known data structure  for approximate membership -- is an HV where the superposition operation is obtained by the component-wise disjunction of binary HVs of symbols (as in SBDR) and, thus, can be seen as a special case of HDC/VSA~\cite{KleykoABF2020}.

Also note that, in principle, multiplicative bindings can be used to represent sets~\cite{MitrokhinSensorimotor2019, KleykoCommentariesSR2020}, however, the similarity properties will be completely different from the representation by the superposition.

\subsubsection{Role-filler bindings} 
 \label{sec:key:value}

Role-filler bindings, which are also called slot-filler or key-value pairs, are a very general way of representing structured data records. For example, in localist representations the role (key) is the component's ID, whereas the filler (value) is the component's value. 
In HDC/VSA, this is represented by the result of binding. Both multiplicative binding and binding by permutation can be used. With multiplicative binding, both the role and the filler are transformed to HVs, which are bound together. When using permutation, it is common that the filler is represented by its HV, while the role is associated with either a unique permutation or a number of times that some fixed permutation should be applied to the filler's HV.

In the case of multiplicative binding, the associations do not have to be limited to two HVs and, therefore, data structures involving more associations can be represented. For example, the representation of ``schemas''~\cite{NeubertHDLearnPlan2018} might involve binding three HVs corresponding to \{\textit{context}, \textit{action}, \textit{result}\}\footnote{Note that the original proposal in~\cite{NeubertHDLearnPlan2018} was based of the superposition of three  role-filler bindings.}. 
It is worth mentioning that in symbolic role-filler bindings roles are usually atomic symbols. In HDC/VSA, role HVs are not required to be representations of atomic objects. They may represent compositional structures and may be constructed to have a desired similarity between roles.

A set of role-filler bindings is represented by the superposition of their HVs. Note that the number of (role-filler) HVs superimposed is limited to preserve the information contained in them, e.g., if one wants to recover the  HVs being superimposed (see Sections~\ref{sec:recovery} and~\ref{sec:capacity}).

\subsection{Numeric scalars and vectors}
\label{sec:scalars:vectors}

In practice, numeric scalars are the data type present in many tasks, especially as components of vectors. 
Representing close numeric values with i.i.d. random HVs does not preserve the similarity of the original data. 
Therefore,  when transforming numeric data to HVs, the usual requirement is that the HVs of nearby data points are similar and those of distant ones are dissimilar. 

We distinguish three approaches to the transformation of numeric vectors to HVs~\cite{RachkovskijWorldModel2013}: compositional (Section~\ref{sec:scalars:vectors:compos}), explicit receptive fields (Section~\ref{sec:RecFields}), and random projections (Section~\ref{sect:rand:proj}). 

\subsubsection{Compositional approach}
\label{sec:scalars:vectors:compos}

\paragraph{Representation of scalars in the compositional approach}
In order to represent close values of a scalar by similar HVs, correlated HVs should be generated, so that the similarity decreases with the increasing difference of scalar values. 
Usually, a scalar is first normalized into some pre-specified range (e.g., $[0,1]$). Next it is quantized into finite grades (levels). Since HVs have a finite dimension and are random, they can reliably (with non-negligible differences in similarity) represent only a finite number of scalar grades. So only a limited number of grades are represented by correlated HVs, usually up to several dozens.

Early schemes for representing scalars by HVs were proposed independently in~\cite{RachkovskijAudio1990} and~\cite{SmithScatter1990}. 
These and other schemes~\cite{RachkovskijScalars2005, PurdyEncoding2016}
can be considered as implementing some kind of ``flow'' from one component set of an HV to another, with or without substitution. 
For example, ``encoding by concatenation''~\cite{RachkovskijScalars2005} uses two random HVs: one is assigned to the minimum grade in the range, while the second one is used to represent the maximum grade. The HVs for intermediate grades are formed using some form of interpolation, such as the concatenation of the parts of the HVs where the lengths of these two parts  are
proportional to the distances from the current grade to the minimum and maximum one. 
Note that in this scheme the representations of all values in the range have some nonzero degree of similarity to each other (apart from the two extreme values). It is possible to apply the same scheme to multiple concatenated ranges, so that the values separated by more than the range size have zero similarity.

Similar types of interpolations were proposed in~\cite{CohenOrthogonality2013, WiddowsContinuous2015, KleykoTradeoffs2018}. 
The scheme in~\cite{SmithScatter1990} and 
the ``subtractive-additive'' one~\cite{RachkovskijScalars2005} 
start from a random HV and recursively flip the values of some components to obtain the HVs of the following grades. This scheme was  popularized, e.g.,  in~\cite{RahimiBiosignal2016,KleykoTradeoffs2018}, for representing values of features when applying HDC/VSA in classification tasks.
Another early scheme proposed for complex and real-valued HVs is called fractional power encoding (see Section 5.6 in~\cite{PlateNested1994}). 
It is based on the fact that complex-valued HVs (random angles on a unit circle) can be (component-wise) exponentiated to any value:  
\noindent
\begin{equation}
\mathbf{z}(x)=\mathbf{z}^{\beta x},
\label{eq:fpe:scalar}
\end{equation}
\noindent
where $\mathbf{z}(x)$ is an HV representing scalar $x$, $\mathbf{z}$ is a random HV called the base HV~\cite{FradyFunctions2021, FradyFunctionsNICE2022}, and $\beta$ is the bandwidth parameter controlling the width of the resultant similarity kernel.
This approach  requires neither normalizing scalars in a pre-specified range nor quantizing them. 
The fractional power encoding can be immediately applied to FHRR and to any other representation where the phase angle resolution is sufficient. 
It is also used for HRR by making complex-valued HV using fast Fourier transform, exponentiating, and then making inverse fast Fourier transform to return to real-valued HVs (see~\cite{FradyFunctions2021, FradyFunctionsNICE2022} for the details).

\paragraph{Representation of numeric vectors in the compositional approach}
In the compositional approach to numeric vector representation by HVs (~\cite{RachkovskijAudio1990, StanfordScatter1994, RachkovskijVectors2005}, we first form HVs for scalars (i.e., for components of a numeric vector).
Then, HVs corresponding to values of different components are combined (often via superposition but the multiplicative binding is used too) to form the compositional HV of the numeric vector.

When constructing the compositional HV, it is important to make sure that scalar values of different components are represented by dissimilar HVs (even when two components have the same value).  
Otherwise, the association between a component and its value is lost. 
If scalars in different components are represented by dissimilar HVs, they can be superimposed or bound as is.     
Sometimes it is more practical (e.g., to reduce memory requirements) to keep a single set of correlated HVs representing all considered values of scalars~\footnote{
This set of HVs is sometimes called ``continuous item memory'' since HVs are correlated.}. 
The set is shared across all components.
In this case, prior to applying the superposition operation,  one has to associate the scalar's HV with its component identity in the numeric vector.
This can be done by representing role-filler bindings (see Section~\ref{sec:key:value}) with some form of binding. 
This means that when using the role-filler binding scheme, the HV of a scalar corresponds to the filler while its component identity in the numeric vector corresponds to the role.
As presented in Section~\ref{sec:key:value}, the association between role's and filler's HV can be realized either via the multiplicative binding with a role's HV or by permuting a filler's HV using the permutation assigned to the role.    

Note that different schemes and models provide 
different types of similarity, for example, Manhattan distance ($\text{dist}_{\text{Man}}$) in~\cite{RachkovskijVectors2005, ThomasHDFoundations2020}.
Note also that, unlike usual vectors, where the components are orthogonal, role HVs of nearby components could be made to correlate (e.g., nearby filtration banks are more correlated than banks located far away)~\cite{RachkovskijAudio1990}.

\subsubsection{Explicit receptive fields}
\label{sec:RecFields}

The receptive field approach is often called coarse coding since a numeric vector is coarsely represented by the receptive fields activated by the vector. There are several well-known schemes utilizing explicit receptive fields, such as Cerebellar Model Articulation Controller (CMAC)~\cite{AlbusCMAC1975}, Prager Codes~\cite{PragerCodes1993}, and 
Random Subspace Codes (RSC)~\cite{KussulRandom1999, KussulThreshold1994}, where receptive fields are various kinds of randomly placed and sized hyperrectangles, often in only some of the input space dimensions. It is worth noting that these schemes form (sparse) binary HVs. Similar approaches, however, can produce real-valued HVs, e.g., by using Radial Basis Functions as receptive fields~\cite{AisermanPotential1964, MoodyFast1989}.
Details and comparisons can be found in~\cite{RachkovskiiRropertiesRSC2005, RachkovskiiResolutionRCS2005}. In particular, a similarity function between the input numeric vectors represented by RSCs was obtained.

\subsubsection{Random projections-based approach}
\label{sect:rand:proj}

Random projection (RP) is another approach which allows forming an HV of a numeric vector $\mathbf{x}$ by multiplying it by an RP matrix $\mathbf{R}$: 
\noindent
\begin{equation}
\mathbf{z}= \mathbf{R} \mathbf{x},
\label{eq:RP}
\end{equation}
\noindent
and possibly performing some normalization, binarization and/or sparsification (such as preserving $k$ largest vector components).\footnote{ 
Note that (\ref{eq:RP}) represents the values of $\mathbf{x}$ via the magnitudes of the corresponding columns in $\mathbf{R}$, so that $\mathbf{z}$ is the weighted superposition of the columns in $\mathbf{R}$. 
This approach to representing numeric vectors is different from the ideas in Section~\ref{sec:scalars:vectors:compos} above, where the difference in the direction of HVs is used to represent relative similarity between the corresponding scalars.  
However, the representation of numeric vectors via weighted superposition is rarely used in HDC/VSA since the result of (\ref{eq:RP}) needs to be normalized, and, thus, the information about the absolute magnitudes of $\mathbf{z}$ is lost (see Section~\ref{sec:vsa:operations:superposition}).
} 
It is a well-known approach, which has been extensively studied in mathematics and computer science~\cite{JohnsonJL1984, IndykApproximate1998, PapadimitriouLatent2000, VempalaRP2005, RachkovskijRPBook2018}. Originally, the RP approach was used in the regime where the resultant vectors $\mathbf{R} \mathbf{x}$ of smaller dimensionality were produced (i.e., used for dimensionality reduction). 
This is useful for, e.g., fast estimation of $\text{dist}_{\text{Euc}}$ of original high-dimensional numeric vectors (as well as $\text{sim}_{\text{dot}}$ and $\text{dist}_{\text{angle}}$). 
Well-known applications are in similarity search~\cite{KaskiDimensionality1998, IndykApproximate1998} and compressed sensing~\cite{DonohoCompressed2006}. 
First, a random $\mathbf{R}$ with components from the normal distribution was investigated~\cite{JohnsonJL1984, IndykApproximate1998, VempalaRP2005}, but then the same properties were proved for RP matrices with components from $\{-1,+1\}$ (bipolar matrices) and $\{-1,0,+1\}$ (ternary and sparse ternary matrices)~\cite{AchlioptasFriendlyRP2003, LiSparseRP2006,KaneSparser2014}.

In the context of HDC/VSA, the RP approach using a sparse ternary matrix was first applied in~\cite{KanervaRI2000} (see also Section~\ref{PartII-sec:random:indexing} of Part~II of the survey~\cite{KleykoSurveyVSA2021Part2} for details). 
In~\cite{MisunoSearching2005,MisunoVector2005} (see Section~\ref{PartII-sec:context:HVs} of Part~II of the survey~\cite{KleykoSurveyVSA2021Part2}), it was proposed to binarize or ternarize (by thresholding) the result of $\mathbf{R} \mathbf{x}$, which produced sparse HVs. 
The analysis of a $\text{sim}_{\text{cos}} $  estimation by sparse HVs was first reported in 2006/2007 and then published in~\cite{RachkovskijRP2012, RachkovskijBinaryRP2014, RachkovskijSimilarityRP2015}, whereby the RP with sparse ternary and binary matrices was used. 
In~\cite{DasguptaNeural2017}, the RP with sparse binary matrices was used for expanding the dimensionality of the original vector. 
Note that increasing the dimensionality of the original vector has been widely used previously in machine learning classification tasks as it allows linearizing nonlinear class boundaries. In~\cite{BeckerNNSearch2016}, thresholding of the HV produced by $\mathbf{R} \mathbf{x}$  was investigated and applied in the context of fast (i.e., sublinear versus the number of objects in the base) similarity search. 
It is common to use a single RP matrix to form an HV, but there are some approaches that rely on the use of several RP matrices~\cite{RasanenMultivariate2015}:
$\mathbf{z}= \sum_{i=1}^{n} \lambda_i \mathbf{R}_i \mathbf{x}$,
where $\lambda_i$ determines the contribution of the $i$th RP $\mathbf{R}_i \mathbf{x}$ to the resultant HV and, thus, can be interpreted as a weighted superposition of $n$ separate representations.

An appealing feature of the RP approach is that it can be applied to input vectors of an arbitrary dimensionality and number of component gradations, unlike the compositional or the receptive fields approaches. 
Some theoretical analysis of using RP for forming HVs was performed in~\cite{ThomasHDFoundations2020}.

\subsection{Sequences}
\label{sec:sequences}

Below we present several standard approaches for representing sequences with HVs. These approaches include  binding of elements' HVs with their position HVs,  bindings using permutations, binding with the context HVs, and using HVs for $n$-grams. For an overview of these approaches refer to, e.g.,~\cite{SokolovSequence2006}.

\subsubsection{Binding of elements' HVs with their position or context HVs}

One of the standard ways of representing a sequence is to use a superposition of role-filler binding HVs where the filler is a symbol's HV and the role is its position's HV~\cite{KussulSequences1991, PlateSequences1995, HannaganHolographic2011}. 
In \cite{KussulSequences1991}, the order was represented by the weight of the subset of components selected from the initial HVs of sequence symbols (e.g., each preceding symbol was represented by more 1-components of the binary HV than the succeeding symbol). In \cite{RachkovskijPhD1990}, it was proposed to use thinning from the HVs of the previous sequence symbols (context).
It is, however, most common to use i.i.d. random HVs to represent positions.
An approach to avoid storing HVs for different positions is to use HDC/VSA models where the binding operation is not self-inverse. In this case, the absolute position  $k$ in a sequence is represented by binding the position HV to itself $k$ times ~\cite{PlateRecurrent1992, PlateSequences1995}. 
Such a mechanism is called ``trajectory association''. 
The trajectory association can be convenient when the sequence representation is formed sequentially,  allowing the sequence HV to be formed in an incremental manner. However, the sequence HV does not preserve the similarity to symbols' HVs in nearby positions, since the position HVs are not similar and so the role-filler binding HVs are also dissimilar too. The MBAT model (Section~\ref{sec:framework:MBAT}) has similar properties, where the position binding is performed by multiplying by a random position matrix.

In order to make symbols in nearby positions similar to each other, one can use correlated position HVs~\cite{PlateNested1994, SokolovSequence2006} such that, e.g., shifted sequences will still result in a high similarity between their HVs. This approach was also proposed independently in~\cite{CohenOrthogonality2013,WiddowsContinuous2015} (see Section~\ref{sec:scalars:vectors}).
Similar ideas for 2D images are discussed in Section~\ref{sec:2Dimages:role-filler}.

Also, the HV of the next symbol can be bound to the HV of the previous one, or to several (weighted) HVs of previous ones, i.e., implementing the ``binding with the context'' approach \cite{RachkovskijPhD1990}.

\subsubsection{Representing positions via permutations}
\label{sec:sequences:permutations}

An alternative way of representing position in a sequence is to use permutation~\cite{kussul1993information, SahlgrenOrder2008, KanervaHyperdimensional2009}. Before combining the HVs of sequence symbols, the order $i$ of each symbol is represented  
by applying some specific permutation to its HV $i$ times (e.g., $\rho^2(\mathbf{c})$). 
However, such a permutation-based representation does not preserve the similarity of the same symbols in nearby positions, since the permutation does not preserve the similarity between the permuted HV and the original one. 
To preserve the similarity of HVs when using permutations, in~\cite{KussulPermutative2003, CohenEARP2018}, it was proposed to use partial (correlated) permutations.

\subsubsection{Compositional HVs of sequences}

Once symbols of a sequence are associated with their positions, the last step is to combine the sequence symbols' HVs into a single compositional HV representing the whole sequence. There are two common ways to combine these HVs. 

The first way is to use the superposition operation, similar to the case of sets in Section~\ref{sec:sets}. For example, for the sequence (a,b,c,d,e) the resultant HV (denoted as $\mathbf{s}$) is: 
\noindent
\begin{equation}
\mathbf{s} = \rho^0(\mathbf{a}) + \rho^1(\mathbf{b}) + \rho^2(\mathbf{c}) + \rho^3(\mathbf{d}) + \rho^4(\mathbf{e}). 
\end{equation}
\noindent
Here we exemplified the representation of positions via permutations but the composition will be the same for other approaches. 
The advantage of the approach with the superposition operation is that it is possible to estimate the similarity of two sequences by measuring the similarity of their HVs.

The second way of forming the compositional HV of a sequence involves binding of the permuted HVs, e.g., the sequence above is represented as (denoted as $\mathbf{p}$): 
\noindent
\begin{equation}
\mathbf{p} = \rho^0(\mathbf{a})  \circ \rho^1(\mathbf{b})  \circ \rho^2(\mathbf{c})  \circ \rho^3(\mathbf{d})  \circ \rho^4(\mathbf{e}).  
\end{equation}
The advantage of this sequence representation is that it allows forming quasi-orthogonal HVs even for sequences that differ in only one position.
Similar to the trajectory association, extending a sequence can be done by permuting the current sequence's HV and adding or multiplying it with the next HV in the sequence, hence, incurring a fixed computational cost per symbol.

Note that compositional HVs of sequences can also be of a hybrid nature. 
For example, when representing positions via permutations the similarity of the same symbols in nearby positions is not preserved. 
One way to preserve similarity when some of the symbols in a sequence are permuted was presented in \cite{KleykoPermuted2016}, where compositional HVs of sequences
included two parts: a representation of a multiset of symbols constituting a sequence (a bag-of-symbols) and an ordered representation of symbols. 
The first part is transformed into an HV as a multiset of symbols (Section~\ref{sec:sets}) by superimposing atomic HVs of all symbols present in a sequence. 
The second part is transformed into an HV as a sequence using permutations of symbols' HVs to encode their order (Section~\ref{sec:sequences:permutations}). 
Both representations are superimposed together to obtain the compositional HV.

In~\cite{RachkovskijEquivariant2021, RachkovskijRecursiveBinding2022}, it was proposed to form such representations of sequences that preserve similarity in nearby positions as well as being equivariant with respect to sequence shifts. The property of equivariance means that the HV of the shifted sequence can be obtained not only by transforming this sequence into an HV, but just by some transformation of the HV of the original sequence. 
To achieve these properties of representations,~\cite{RachkovskijEquivariant2021} formed symbols' HVs using the superposition of several appropriately permuted versions of the corresponding atomic HV. 
The same idea was also used in~\cite{RachkovskijRecursiveBinding2022} but with recursive applications of multiplicative binding instead of permutations. 

\subsubsection{\textit{n}-grams}
\label{sec:sequences:ngrams}

$n$-grams are $n$ consecutive symbols of a sequence.
In the $n$-gram representation of a sequence, all $n$-grams of the sequence are extracted, sometimes for different $n$. Usually, a vector containing $n$-gram statistics is formed such that its components correspond to different $n$-grams. The value of the component is the frequency (counter) of the occurrence of the corresponding $n$-gram.

There are various transformations of $n$-grams into HVs. 
The possibility of representing multisets with HVs (Section~\ref{sec:sets}) allows forming HVs representing $n$-gram statistics as the superposition of HVs of individual $n$-grams.
So the representations of $n$-grams in HVs are usually different in the following aspect: 
\begin{itemize}
    \item They distinguish different $n$-grams, where ``different'' means even a single unique symbol (e.g., (abc) vs. (abd)) or the same symbols in different positions (e.g., (abc) vs. (acb));
    \item They form similar HVs for similar $n$-grams. 
\end{itemize}

To form dissimilar HVs for different $n$-grams, each $n$-gram can be assigned an i.i.d. random HV. 
However, an $n$-gram HV is usually generated from the HVs of its symbols. 
So in order to save space for the HV representations of $n$-grams, it is possible to represent them using compositional HVs \cite{RachkovskijApplication1996, JoshiNgrams2016} instead of generating an atomic HV for each $n$-gram.
Thus, the approaches above for representing sequences can be used. 
In~\cite{RachkovskijApplication1996, JoshiNgrams2016, KimEfficientDNA2020}, the permuted HVs of the $n$-gram symbols were bound by multiplicative binding.
Alternatively, in~\cite{JonesMeaning2007, KachergisOrBEAGLE2011}, the multiplicative binding of pairs of HVs was performed recursively, whereby the left HV corresponds to the already represented part of the $n$-gram, and the right HV corresponds to the next element of the $n$-gram. 
The left and right HVs are permuted by different random permutations.

In contrast to the approaches above, in~\cite{RecchiaSemantic2010, ImaniDNA2018} the permuted HVs of the $n$-gram symbols were not bound multiplicatively, but were superimposed forming similar HVs for similar $n$-grams. 
In~\cite{HannaganHolographic2011}, the superposition of HVs corresponding to symbols in their positions was also used. However, the position of a symbol in a bi-gram (possibly from non-adjacent symbols) was specified by the multiplicative binding with the HVs of the left and right positions instead of permutations.

\subsubsection{Stacks}
\label{sec:sequences:stacks}

A stack can be seen as a special case of a sequence where elements are either inserted or removed in a last-in-first-out manner. At any given moment, only the top-most element of the stack can be accessed and elements written to the stack before are inaccessible until all later elements are removed. 
HDC/VSA-based representations of stacks were proposed in~\cite{PlateNested1994, StewartSentenceProcessing2014,YerxaUCBHD_FSA2018}. 
The HV of a stack is essentially the representation of a sequence with the superposition operation that always moves the top-most element to the beginning of the sequence.

\subsection{2D images}
\label{sec:2Dimages}

There are a number of proposals for representing objects with a two-dimensional structure, such as 2D images, in HDC/VSA models. In this section, we cluster the existing proposals for representations into three groups\footnote{
There are proposals that cannot be placed unambiguously in one of the clusters. 
For example, the proposal from~\cite{KellyStructure2013} directly used  pixels values that were first flattened and normalized and then shuffled by some random permutation.
The resultant vector was bound to itself with circular convolution to obtain a ``trace'' HV.  
Thus, this proposal is neither purely ``permutation-based'' nor ``role-filler binding-based'' as it relies on both.
Nevertheless, the identified groups are instructive as they are used as design primitives in many transformations of 2D images into HVs.   
}: permutation-based, role-filler binding-based, and neural network-based.

\subsubsection{Permutation-based representations}
\label{sec:2D:permute}

As discussed in Section~\ref{sec:binding}, permutations are used commonly as a way of implementing binding with the position information. 
In the context of 2D images, permutations can be used by assigning one permutation for the $x$-axis and another one for the $y$-axis.

The simplest way of implementing a permutation is to use a cyclic shift as in~\cite{kussul1993information, KussulNNMM2010}.
In order to implement two orthogonal shifts corresponding to the $x$- and $y$-axes, the HVs of features were restructured as a 3D tensor with two dimensions corresponding to the retina size along the $x$- and $y$-axes. Then the HV of a feature at some $(x,y)$ position was obtained by shifting the corresponding 3D tensor $x$ times along the $x$-axis and $y$ times along the $y$-axis. 

%
In a similar way, when using permutations as shift generalizations, 
to bind an HV of a pixel's value (or of an image feature) with its position, $x$-axis permutation is applied $x$ times and $y$-axis permutation is applied $y$ times~\cite{MitrokhinSensorimotor2019, KleykoCommentariesSR2020}. 
The image is represented as a compositional HV containing the superposition of all permuted HVs of pixels' values. 
HVs representing pixels' (features') values are formed some approach for representing scalars. 
As permuted HVs are not similar in any of the works above, the same pixel's value even in nearby positions will be represented by dissimilar HVs.

The issue of the absence of similarity in permutation-based representations was addressed in~\cite{KussulPermutative2003, KussulPermutation2006} by using partial permutations. The number of permuted components of some pixel's value HV (or, more generally, of some feature's HV) increases with the coordinate change (within some radius), up to the full permutation. So, inside the radius the similarity decreases. Outside the radius, the same process is repeated again. As a result, a pixel's value HV at (\textit{x},\textit{y}) is similar to the HVs in nearby positions within the similarity radius. 
The features at positions were Random Local Descriptors (RLD) detected at some points of interest. 
For instance, for binary images, the RLD feature is detected if some (varied for different features) pixels inside a local receptive field take zero and one values. Note that RLD may be considered as a version of receptive fields approach (Section~\ref{sec:RecFields}) and can form HVs representing images, as in the case of the LInear Receptive Area (LIRA) features ~\cite{KussulLiRA2004}.

\subsubsection{Role-filler binding-based representations}
\label{sec:2Dimages:role-filler}

In a 2D image, it is natural to represent a pixel's position and its value as a role-filler binding, where the pixel's position HV is the role and an HV corresponding to the pixel's value is the filler. 
Similar to the permutations, the image HV is formed as a compositional HV containing the superposition of all role-filler binding HVs. 

When it comes to pixels position HVs, the simplest approach is to treat each position as if it would be a unique symbol so that it can be assigned with a unique random HV. 
This approach allows forming image HVs through the role-filler bindings.
It was investigated in~\cite{KleykoHolographic2017, ManabatCharacter2019} for real-valued and dense binary HVs and in~\cite{KleykoSDR2016} for sparse binary HVs. 
The approach with unique HVs for roles, however, neither imposes any relations between pixels' position HVs nor does it allow preserving the similarity of pixels' position HVs in a local neighborhood, which, e.g., is useful for robustness against small changes in the 2D image. 
To address the latter issue, in~\cite{KussulSmallImages1992} a method was described where the HVs of nearby $x$ and $y$ coordinates were also represented as correlated HVs, using approaches for representing scalars (Section~\ref{sec:scalars:vectors}). Then the pixel value in a position (\textit{x},\textit{y}) was represented by binding three HVs: one representing a pixel's value at (\textit{x},\textit{y}) and two representing the $x$ and $y$ coordinates themselves (component-wise conjunction was used for the binding in~\cite{KussulSmallImages1992}). Thus, similar values in close proximity were represented by similar HVs. 

MOving MBAT (MOMBAT)~\cite{GallantPositional2016} was proposed to explicitly preserve the similarity of HVs for nearby pixels. As in the MBAT model (Section~\ref{sec:framework:MBAT}), matrices were used as roles. In particular, the matrix for each coordinate (\textit{x},\textit{y}) was formed as a weighted superposition of two random matrices. Thus, for both x and y, nearby coordinates were represented by similar (correlated) matrices. Then, each (\textit{x},\textit{y}) pair was represented by a matrix obtained by binding (multiplying) those corresponding matrices. A value of a pixel at (\textit{x},\textit{y}) was represented by binding its HV with the matrix of (\textit{x},\textit{y}). Finally, the obtained HVs were superimposed. All the HVs participating in MOMBAT had real-valued components. A similar approach with local similarity preservation for BSC and dense binary HVs was considered in~\cite{BurcoNeuralSymbolic2018}.

The Fractional Power Encoding approach (Section~\ref{sec:scalars:vectors:compos}) to the representation of 2D images was employed in~\cite{WeissOlshausenSpatial16, FradyDisentangling2018, KomerContinuous2019}.
It imposes a structure between pixels' positions and can preserve similarity in a local neighborhood. 
In particular, the image HV is formed as follows. Two random base HVs are assigned for the $x$-axis and $y$-axis, respectively. In order to represent $x$ and $y$ coordinates, the corresponding base HVs are exponentiated to coordinate values; and $\mathbf{z}(x,y)$ - the HV for the (\textit{x},\textit{y}) coordinate - is formed as their binding (assuming $\beta=1$ cf.~(\ref{eq:fpe:scalar})): 
\noindent
\begin{equation}
\mathbf{z}(x,y)=\mathbf{x}^x \circ \mathbf{y}^y.
\end{equation}
\noindent
Finally, the binding with the HV of the pixel value is done. The variant of the fractional power encoding-based representation in hexagonal coordinates was presented in~\cite{KomerNavigation2020}.

\subsubsection{Neural network-based representations}
\label{sec:2Dimages:neural:nets}

Since  neural networks are currently one of the main tools for processing 2D images, it is becoming common to use them for producing HVs of 2D images. 
This is especially relevant since directly transforming pixel values into HVs does not usually provide a good representation for solving machine learning tasks with competitive performance.
For example, there were some studies~\cite{ManabatCharacter2019, hassan2021hyper} directly transforming 2D images in MNIST to HVs. 
Their accuracy was further improved by either enhancing the HV encoder (e.g., with a receptive field and max pooling layer~\cite{DistriHD2022}, or learning binary class HVs from a cross-entropy loss~\cite{LeHDC2022}. 
However, the reported accuracy was, in fact, lower than the one obtained with, e.g., a $k$NN classifier applied directly to pixels' values, or a simple neural network with two layers.
Recall, however, that the HV representations must be designed to capture the information that is important for solving the problem and presenting it in a form that can be exploited by the HDC/VSA mechanisms.
Therefore, it is important to either apply some feature extraction techniques (as was, e.g., the case in~\cite{KussulLiRA2004,KussulPermutation2006}) or use neural networks as a front-end. One of the earliest attempts to demonstrate this was presented in~\cite{YilmazSymbolic2015, YilmazMachine2015}. 
The proposed approach takes activations of one of the hidden layers of a convolutional neural network caused by an input image. 
These activations are then binarized and the result of the binarization is used as an initial grid state of an elementary cellular automaton. 
The automaton is evolved for several steps and the results of previous steps are concatenated together, the result of which is treated as an HV corresponding to the image. 
The automaton evolution performs nonlinear feature extraction, in the spirit of LIRA and RLD.

Later works used activations of convolutional neural networks without extra cellular automata computations. For example, in~\cite{ NeubertRobotics2019, MitrokhinCNN2020} pre-trained off-the-shelf convolutional neural networks were used, while in~\cite{KarunaratneHDAugmented2021,HerscheContinualLearn2022} a network's attention mechanism and loss function were specifically designed to produce HVs with desirable properties.
For example, it directed the output of the neural network to assign quasi-orthogonal HVs to images from different classes~\cite{KarunaratneHDAugmented2021}.
This quasi-orthogonality of the class HVs also enables continual learning of many novel classes, from very few training samples, with small interference, which leads to the state-of-the art classification accuracy on natural and handwritten images~\cite{HerscheContinualLearn2022}.  
It is important to note that obtaining HVs from neural networks is a rather new direction and there are no studies that would scrupulously compare HVs obtained from different neural network architectures.

\subsection{Graphs}
\label{sect:graphs}

\subsubsection{Undirected and directed graphs}
\label{sect:graphs:undirected}

\begin{figure}[tb]
\begin{center}
\begin{minipage}[h]{0.59\linewidth}
\centering
\includegraphics[width=1.0\linewidth]{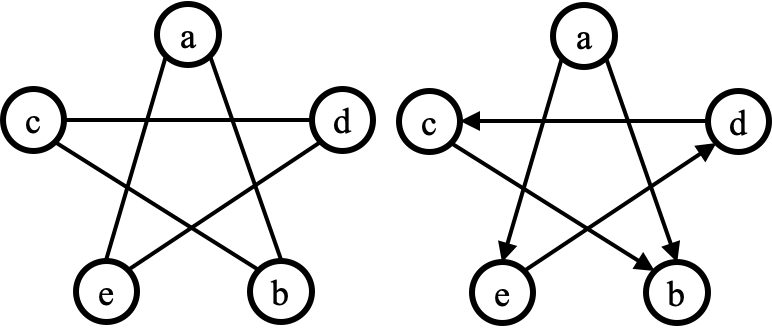}
\caption{
An example of an undirected and a directed graph with $5$ nodes. In the case of the undirected graph, each node has two edges. 
}
\label{fig:graph}
\end{minipage}
\hfill 
\begin{minipage}[h]{0.4\linewidth}
\centering
\includegraphics[width=0.7\columnwidth]{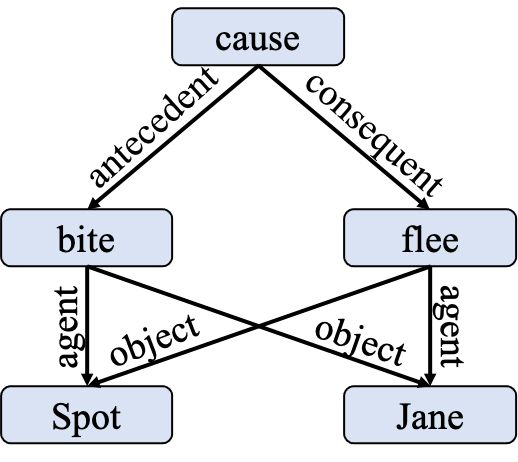}
\caption{
An example of a labelled graph representing the episode -- \texttt{cause(bite(Spot,Jane),flee(Jane,Spot))}.
}
\label{fig:graph:episode}
\end{minipage}
\end{center}
\vspace{-0.3cm}
\end{figure}

A graph, denoted as $G$, consists of nodes and edges. Edges can either be undirected or directed. Fig.~\ref{fig:graph} presents examples of both directed and undirected graphs. 
First, we consider the following simple transformation of graphs into HVs~\cite{GaylerIsomorphism2009}. A random HV is assigned to each node of the graph, following Fig.~\ref{fig:graph} node HVs are denoted by letters (i.e., $\mathbf{a}$ for node ``a'' and so on).
An edge is represented as the binding of HVs of the connected nodes, e.g., the edge between nodes ``a'' and ``b'' is $\mathbf{a} \circ \mathbf{b}$. The whole graph  is represented simply as the superposition   of HVs (denoted as $\mathbf{g}$) representing all edges in the graph, e.g., the undirected graph in Fig.~\ref{fig:graph} is: 
\noindent
\begin{equation}
\mathbf{g} = \mathbf{a} \circ \mathbf{b} + \mathbf{a} \circ \mathbf{e} + \mathbf{b} \circ \mathbf{c} + \mathbf{c} \circ \mathbf{d} + \mathbf{d} \circ \mathbf{e}.  
\end{equation} 
\noindent

To represent  directed graphs, the directions of the edges should be included into their HVs. This could be done, e.g., by applying a permutation, so that the directed edge from node ``a'' to ``b'' in Fig.~\ref{fig:graph} is represented as $ \mathbf{a} \circ \rho(\mathbf{b})$.
Thus, the directed graph in Fig.~\ref{fig:graph} is represented by the following HV:
\noindent
\begin{equation}
\mathbf{g} =
\mathbf{a} \circ \rho(\mathbf{b}) + 
\mathbf{a} \circ \rho(\mathbf{e}) + 
\mathbf{c} \circ \rho(\mathbf{b}) + 
\mathbf{d} \circ \rho(\mathbf{c}) + 
\mathbf{e} \circ \rho(\mathbf{d}).
\end{equation}
\noindent
The elements of the graph $\mathbf{g}$ can be recovered using an item memory (Section~\ref{sec:recovery}).
For graphs that have the same node HVs, the dot product is a measure of the number of overlapping edges. 
The described graph representations do not represent isolated vertices, but this could be fixed, 
by, e.g., separate representations of the vertex set and edge set~\cite{GaylerIsomorphism2009}.

\subsubsection{Labelled directed ordered graphs}
\label{sect:graphs:labelled}

When modeling, e.g., analogical reasoning (Section~\ref{PartII-sec:analogical:reasoning} of Part~II of the survey~\cite{KleykoSurveyVSA2021Part2}) or knowledge graphs, it is common to use graphs where both edges and nodes have associated labels, and edges are directed. Let us consider the representation of a graph shown in Fig.~\ref{fig:graph:episode}, which can also be written in the bracketed notation as: \texttt{cause(bite(Spot,Jane),flee(Jane,Spot))}.

Using the role-filler binding approach (see Section~\ref{sec:key:value}), a relation, e.g., \texttt{bite(Spot,Jane)}, can be represented by HVs (in HRR or BSC) as follows~\cite{PlateHolographic2003}:
\noindent
\begin{equation}
\mathbf{BITE}= \langle\mathbf{bite} + \langle\mathbf{spot} + \mathbf{jane}\rangle + \mathbf{bite}_{\text{agent}} \circ  \mathbf{spot} 
+ \mathbf{bite}_{\text{object}} \circ  \mathbf{jane}\rangle.
\end{equation}
\noindent
Here $\mathbf{bite}$, $\mathbf{bite}_{\text{agent}}$, and $\mathbf{bite}_{\text{object}}$ are atomic HVs, while $\mathbf{spot} = \langle \mathbf{spotID}  + \mathbf{dog} \rangle$ and $\mathbf{jane} = \langle \mathbf{janeID}  + \mathbf{human} \rangle$  are the compositional ones. 
In the same way, $\mathbf{FLEE}$ is formed, and finally the HV of the whole graph: 
\noindent
\begin{equation}
\langle\mathbf{cause} + \langle\mathbf{BITE} + \mathbf{FLEE}\rangle + \mathbf{cause}_{\text{antecedent}} \circ  \mathbf{BITE} 
+ \mathbf{cause}_{\text{consequent}} \circ  \mathbf{FLEE}\rangle.
\end{equation}
\noindent
This particular representation was chosen to preserve the similarity of the resultant HV to the HVs of its various elements and influence the similarity of the graph HVs, because of the binding operation properties in HRR and BSC. 
In SBDR, the binding preserves unstructured similarity so the formation of graph's resultant HV is more compact~\cite{RachkovskijStructures2001}: 
\noindent
\begin{equation}
\mathbf{BITE}= 
 \langle  
 \langle \mathbf{bite}_{\text{agent}} \vee  \mathbf{spot} \rangle
 \vee
 \langle \mathbf{bite}_{\text{object}} \vee  \mathbf{jane} \rangle
  \rangle;
\end{equation}
\noindent
\noindent
\begin{equation}
\mathbf{FLEE}= 
 \langle  
 \langle \mathbf{flee}_{\text{agent}} \vee  \mathbf{jane} \rangle
 \vee
 \langle \mathbf{flee}_{\text{object}} \vee  \mathbf{spot} \rangle
  \rangle;
\end{equation}
\noindent
\noindent
\begin{equation}
\mathbf{CAUSE}= 
 \langle  
 \langle \mathbf{cause}_{\text{antecedent}} \vee  \mathbf{BITE} \rangle
 \vee
 \langle \mathbf{cause}_{\text{consequent}} \vee  \mathbf{FLEE} \rangle
  \rangle.
\end{equation}
\noindent
Also, Predicate-Arguments relation representation using random permutations to represent the order of relation arguments was proposed in~\cite{RachkovskijBinding2001, RachkovskijStructures2001, KanervaComputing2019}:
\noindent
\begin{equation}
\mathbf{BITE}= 
 \langle  
 \mathbf{bite} +  \mathbf{spot} +  \mathbf{jane} + \rho_{\text{agent}}(\mathbf{spot}) + \rho_{\text{object}}(\mathbf{jane})
  \rangle;
\end{equation}
\noindent
\noindent
\begin{equation}
\mathbf{FLEE}= 
 \langle  
 \mathbf{flee} + \mathbf{spot} +  \mathbf{jane} + \rho_{\text{agent}}(\mathbf{jane}) + \rho_{\text{object}}(\mathbf{spot})
  \rangle;
\end{equation}
\noindent
\noindent
\begin{equation}
\mathbf{CAUSE}= 
 \langle  
 \mathbf{cause} + \mathbf{BITE} +  \mathbf{FLEE} +
 \rho_{\text{antecedent}}(\mathbf{BITE}) + \rho_{\text{consequent}}(\mathbf{FLEE})
  \rangle.
\end{equation}
\noindent
Note that in the last six equations $\langle \cdot \rangle$ refers to the CDT procedure.
These types of graph representations by HVs will be further discussed in Section~\ref{PartII-sec:analogical:retrieval} of Part~II of the survey~\cite{KleykoSurveyVSA2021Part2} when describing analogical reasoning with HDC/VSA.

Representations of knowledge graphs were further studied  in~\cite{MaHolisticMemoriz2018}. The work proposed to use a Cauchy distribution to generate atomic HVs and demonstrated the state-of-the-art results on the task of inferring missing links using HVs of knowledge graphs as an input to a neural network.   
HVs of knowledge graphs can also be constructed using HVs of nodes and relations that  are learned from data as proposed in \cite{NickelHoloGraph2016}, where HRR was used due to its differentiability.

\subsubsection{Trees}

Trees are an instance of graphs where a node at the lower level (child) belongs to a single higher-level node (parent). Therefore, trees can be represented in the same manner as the graphs above. Please refer to the examples of the transformations to HVs given, e.g., in~\cite{ RachkovskijBinding2001} for unordered binary trees and in~\cite{FradyResonator2020, KleykoComputingParadigm2021} for ordered binary trees.
A representation of trees for the TPR model, using cryptographic hashing algorithms, was investigated in~\cite{HaleyTree2020}.

\section{Discussion}
\label{sec:disc}

In this section, we only discuss of the HDC/VSA aspects covered in this first part of the survey.
Please refer to Part~II~\cite{KleykoSurveyVSA2021Part2} for an extensive discussion of the aspects related to application areas, interplay with neural networks, and open issues.

\subsection{Connections between HDC/VSA models}

As presented in Section~\ref{sec:vsa:frameworks}, there are several HDC/VSA models. Currently, there are many unanswered questions about the connections between models. For example, some models use interactions between multiple components of HVs when performing the multiplicative binding operation (e.g., circular convolution or the Context-Dependent Thinning procedure). Other models (e.g., Fourier Holographic Reduced Representations and Binary Spatter Codes) use multiplicative binding operations that are component-wise. It is not always clear in which situations one is better than the other. 
Therefore, it is important to develop a more principled understanding of the relations between the multiplicative binding operation and of the scope of their applicability. 
For example, a recent theoretical result in this direction~\cite{FradySDR2020} is that under certain conditions the multiplicative binding in Multiply-Add-Permute is mathematically equivalent to the multiplicative binding in Tensor Product Representations.

Moreover, in general it is not clear whether there is one best model, which will dominate all others, or whether each model has its own applicability scope. 
There are works~\cite{KellyStructure2013, SchlegelVSAComparison2020} reporting systematic experimental comparisons between the models for aspects such as, e.g., 
mathematical properties of the operations, information capacity of representations, or robustness to noise. 
We believe that this line of work should be continued since an abstract mathematical characterization of the necessary properties for HDC/VSA might make it easier to generate new HDC/VSA models as alternative instantiations of such properties.
Finally, there is the question of whether there is a need for new HDC/VSA models.

\subsection{Theoretical foundations of HDC/VSA}

It should be noted that HDC/VSA have largely started as an empirical field. At the same time, HDC/VSA implicitly used  well-known mathematical phenomena such as concentration of measure~\cite{FradySDR2020} and random projection~\cite{ThomasHDFoundations2020}.

Currently, there is a demand for laying out solid theoretical principles for HDC/VSA. We are starting to see promising results in this direction. For example, there is a recent ``capacity theory''~\cite{FradyCapacity2018} (Section~\ref{sec:capacity}), which provides a way of estimating the amount of information that could be reconstructed from HVs using the nearest neighbor search in the item memory. Another example is the upper bound guarantees for the reconstruction of data structures from HVs~\cite{ThomasHDFoundations2020}. In~\cite{FradySDR2020} manipulations of HVs in HDC/VSA were shown to be connected to the compressed sensing.

Recently, in~\cite{YuUnderstandingHDC2022}, the expressivity of HDC/VSA models was characterized using similarity matrices. It proved that BSC could not approximate a given similarity matrix arbitrarily well, and could therefore not learn a Bayes optimal classifier. This work has also formalized HDC/VSA using the notion of finite groups and as a result introduced a new model CGR that generalized BSC, and is equivalent to FHRR in the limit.

Another important aspect that should be studied theoretically is the definition of the binding operation. Currently, there is no agreement on what shall be considered a canonical definition of the binding. Various researchers define it differently. For example, is it compulsory that the result of binding is dissimilar to its arguments as is, e.g., the case in FHRR or BSC, or shall it preserve some similarity as in SBDR? An attempt to taxonomize the existing proposals can be found in~\cite{SchlegelVSAComparison2020}, but it does not provide an answer to the question of whether there exists some abstract mathematical definition of a ``pure'' binding operation and whether it is useful in practice to deviate from this definition and create concrete realizations of the binding operation that would not comply completely with that definition.
In that vein, a recent theoretical consideration of multiplicative binding was presented in~\cite{HirataniQuadraticBinding2022}.
%
We foresee that, as HDC/VSA will be exposed more to theoretical computer scientists and applied mathematicians, we will see more works building the theoretical foundations of the field. 

\subsection{Implementation of the HDC/VSA models in hardware}
\label{disc:develop:implmentation}

A large part of the promise of HDC/VSA is based on the fact that they are suitable for implementations on a variety of unconventional hardware such as neuromorphic hardware~\cite{FradyTPAM2019}, in-memory computing~\cite{KarunaratneInMemory2020,KarunaratneHDAugmented2021}, monolithic
3D integration hardware~\cite{Li3DVRRAM2016,WuCarbonNanotube2018}, etc. It is interesting that the motivation to use HDC/VSA on specialized hardware was present from the very beginning. Early examples are specialized neurocomputers~\cite{RachkovskijTexture1991} built to operate with Sparse Binary Distributed Representations. 
Moreover, the topic of hardware implementations has received a lot of attention recently. This is mostly due to the fact that modern machine learning algorithms such as deep neural networks require massive resources to train and run them~\cite{StrubellEnergyNLP}. In our opinion, an additional motivation comes from the fact that HDC/VSA can be seen as an abstraction algorithmic layer and can, thus, be used for designing computational primitives, which can then be mapped to various hardware platforms~\cite{KleykoComputingParadigm2021}. 
As another interesting direction, HDC/VSA can embrace noise and stochasticity of in-memory computing hardware~\cite{KarunaratneInMemory2020,KarunaratneHDAugmented2021,KleykoAugmented2022} to enhance security as shown in~\cite{HyperlockCai2020} by directly generating time-variable cipher-texts that are yet decryptable.
Nevertheless, hardware for HDC/VSA is an active research field and a topic of its own, therefore we have decided to leave it outside the scope of this survey. However, we expect that in the coming years this topic is going to play an important role in the development of the community.

\section{Conclusion}
\label{sec:conc}

In this Part I of the survey, we provided a comprehensive coverage of the computing framework known under the names Hyperdimensional Computing and Vector Symbolic Architectures.
We paid particular attention to existing Hyperdimensional Computing/Vector Symbolic Architectures models and the transformations of input data of various types into hypervector representations.

Part~II of the survey~\cite{KleykoSurveyVSA2021Part2} reviews known applications, touches upon cognitive modeling and cognitive architectures, and discusses the open problems along with the most promising directions for the future work.

\bibliographystyle{apalike}
\bibliography{Bibliography}

\end{document}